\documentclass{article}

    \PassOptionsToPackage{numbers}{natbib}

\usepackage[preprint]{neurips_data_2024}
\usepackage[numbers]{natbib}

\usepackage[utf8]{inputenc} 
\usepackage[T1]{fontenc}    
\usepackage{url}            
\usepackage{booktabs}       
\usepackage{amsfonts}       
\usepackage{nicefrac}       
\usepackage{microtype}      
\usepackage{xcolor}         

\newcommand{\name}[0]{\textsc{Seam}}
\newcommand{\md}[0]{multi-document}
\newcommand{\MD}[0]{Multi-Document}
\newcommand{\Md}[0]{Multi-document}

\newcommand{\asp}[0]{OpenAsp}
\newcommand{\mn}[0]{MultiNews}
\newcommand{\fuse}[0]{FuseReviews}
\newcommand{\musique}[0]{MuSiQue}
\newcommand{\ecb}[0]{ECB+}
\newcommand{\scico}[0]{SciCo}

\newcommand{\llama}[0]{Llama 3}
\newcommand{\mistral}[0]{Mistral}
\newcommand{\mixtral}[0]{Mixtral}
\newcommand{\gemma}[0]{Gemma}

\usepackage[frozencache=true,cachedir=minted-cache]{minted} 
\usepackage{booktabs}
\usepackage{multirow}
\usepackage{graphicx}
\usepackage{caption,subcaption}
\usepackage{wrapfig}
\newcommand{\eqdef}[0]{\stackrel{def}{=}}
\usepackage[utf8]{inputenc}
\usepackage[T1]{fontenc}
\usepackage{mdframed}
\usepackage{todonotes}

\usepackage{lineno}

\usepackage{pifont}
\usepackage[normalem]{ulem}

\title{\name{}: A Stochastic Benchmark \\for Multi-Document Tasks}

\author{%
Gili Lior$^{1,2}$ \quad Avi Caciularu$^{3}$\thanks{Equal contribution.} \quad Arie Cattan$^{3,4*}$ \\ \quad \textbf{Shahar Levy$^{1*}$} \quad   \textbf{Ori Shapira$^{5*}$} \quad \textbf{Gabriel Stanovsky$^{1,2}$} \\
$^1$The Hebrew University of Jerusalem \quad $^2$Allen Institute for AI \\ $^3$Google Research \quad
$^4$Bar-Ilan University \quad $^5$OriginAI}

\begin{document}

\maketitle

\begin{abstract}
Various tasks, such as summarization, multi-hop question answering, or coreference resolution, are naturally phrased over collections of real-world documents. Such tasks present a unique set of challenges, revolving around the lack of coherent narrative structure across documents, which often leads to contradiction, omission, or repetition of information. 
Despite their real-world application and challenging properties, there is currently no benchmark which specifically measures the abilities of large language models (LLMs) on \md{} tasks.
To bridge this gap, we present \name{} (a Stochastic Evaluation Approach for Multi-document tasks), a conglomerate benchmark over a diverse set of \md{} datasets, setting conventional evaluation criteria, input-output formats, and evaluation protocols. In particular, \name{} addresses the sensitivity of LLMs to minor prompt variations through \emph{repeated evaluations}, where in each evaluation we sample uniformly at random the values of arbitrary factors (e.g., the order of documents).
We evaluate different LLMs on \name{}, finding that \md{} tasks pose a significant challenge for LLMs, even for state-of-the-art models with 70B parameters. In addition, we show that the stochastic approach uncovers underlying statistical trends which cannot be observed in a static benchmark.  We hope that \name{} will spur progress via consistent and meaningful evaluation of \md{} tasks.\footnote{\name{} is available at~\url{https://seam-benchmark.github.io}}

\end{abstract}

\section{Introduction}

Many real-world tasks are  performed over document collections \citep{caciularu-etal-2021-cdlm-cross,yasunaga-etal-2022-linkbert}. For example, \md{} summarization requires models to process and integrate information across multiple related documents (e.g., news coverage of an event)~\citep{xiao-etal-2022-primera,caciularu-etal-2023-peek,ernst-etal-2022-proposition}; multi-hop question answering requires models to reason across different pieces of information (e.g., different Wikipedia articles)~\citep{welbl-etal-2018-constructing,tu-etal-2019-multi,caciularu-etal-2022-long}; and cross-document coreference resolution models link entity mentions across documents ~\citep{eirew-etal-2021-wec,eirew-etal-2022-cross,cattan2021scico,cattan-etal-2021-realistic,hirsch-etal-2021-ifacetsum}.

\begin{figure}[t]
\centering
\resizebox{\textwidth}{!}{%
\includegraphics[]{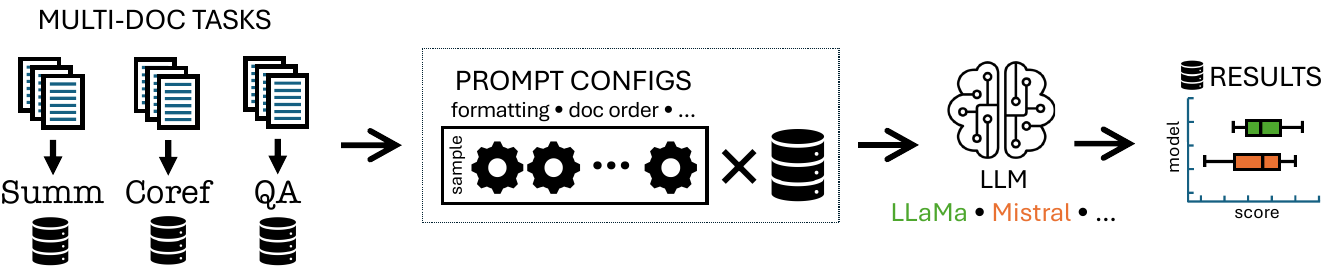}
}
\caption{The \name{} benchmark consists of datasets addressing various multi-document processing tasks. An LLM is  evaluated against different samplings of prompt configurations, producing results that quantify models' performance on \md{} tasks, as well as their  sensitivity to prompt and instance variations.}
\label{fig_overview}
\end{figure}

\Md{} tasks are challenging for several unique aspects compared to single-document applications. First, document collections do not present a coherent narrative. As a result, they are often rife with redundancies, where different authors relate the same information, while also containing many conflicting reports of either factual events (e.g., in the case of an evolving event) or subjective interpretations (e.g., in the case of editorial columns or reviews)~\cite{hirsch-etal-2023-revisiting}. These need to be resolved by the model without any particular source necessarily relating these inter-document relations.
Relatedly, document collections do not have inherent ordering. 
The time of writing may provide a weak signal, however it is not always available, and not always useful, for example, in the case of subjective reports. Finally, \md{} collections exacerbate the challenge of handling long context, as a document collection may contain a large number of documents, each of which may be long on its own.

Despite their real-world motivation and unique properties, there is currently no benchmark which focuses on  the performance of LLMs on multiple fundamental \md{} tasks. Instead, research into \md{} tasks is largely balkanized into separated, disjoint efforts, which do not allow newly built LLMs to easily gauge their \md{} performance and where they currently struggle on such tasks.


In this work, we build \name{} --- the first benchmark geared towards a variety of fundamental \md{} tasks. \name{} measures model performance on samples originating in six existing datasets from three challenging task families: \md{} summarization, multi-hop question answering, and cross-document coreference resolution. In essence \name{} can be seen as a retrieval augmented generation (RAG) setup, where the retrieval is fixed, and the models need to process the retrieved documents.
\name{} is built to benchmark LLM performance, and streamlines the entire evaluation process. From prompt formatting for different tasks, through task-specific post-processing, and standardized evaluation metrics. 



Importantly, \name{} is stochastic by design, and measures model performance across multiple resamples of different arbitrary design choices. Several recent works have noticed that minor and seemingly-arbitrary benchmark variations (e.g., prompt phrasing and formatting) can lead to vastly different results, calling into question results obtained based on a single prompt template~\cite{Perlitz2023EfficientB,sclar2023quantifying,mizrahi2024state,voronov2024mind}.
To cope with this limitation, we design \name{} as a sample \emph{generator} rather than a fixed repository of static task instances. 
We identify five arbitrary benchmarking design choices, including instruction paraphrases, the order of presented documents, the dataset instances used for evaluation, the few-shot examples and their ordering. For each evaluation run, \name{} randomly samples a realization for each of these factors, thus generating a dataset instance,
against which all models are evaluated. 
As illustrated in Figure  \ref{fig_overview}, this enables us a thorough statistical analysis of model performance across a wide range of options and metrics, and hence yields a more reliable performance report across the tasks in our benchmark. 

By testing a wide range of models and configurations, we find that that state-of-the-art models still have room for improvement, and that a larger model size does not guarantee superior mean performance, nor more robust predictions. Further, our analyses show that \md{} tasks present a challenge beyond just the handling of long context, as the the instance length is not correlated with any of the models' success on any of our datasets.

Overall, our contributions are two fold. First, we design, build, and populate the first benchmark specifically designed to test LLM performance on fundamental \md{} tasks. 
\name{} enables 
easy comparison of state-of-the-art LLMs on \md{} tasks, and has the promise to advance the field by standardizing and streamlining its evaluation process. Second, we provide a methodological approach to address the sensitivity of LLMs to seemingly arbitrary choices. Such methodology can be adapted to other datasets and benchmarks, beyond \name{}.


\section{\name{}: A Benchmark for \MD{} Tasks}
\label{sec:dataset}

In this section we describe our data curation for the \name{} benchmark.\footnote{\name{} stands for \emph{a \textbf{S}tochastic \textbf{E}valuation \textbf{A}pproach for \textbf{M}ulti-document tasks}.} \name{} is a conglomeration of samples from existing datasets, specifically designed to satisfy desiderata which together bridge the gap in benchmarking LLMs on \md{} tasks.

First, \name{} is \emph{diverse} in important aspects of the data composition, as it covers a wide range of domains and tasks on real-world texts. Second, \name{} is \emph{challenging and relevant}, for each of our tasks, we choose datasets for which current state-of-the-art performance leaves vast room for improvement. 
Finally, all of the instances in \name{} are phrased using text-only inputs and outputs. This entails linearization of hierarchical structures and post-processing techniques for tasks which require discrete inputs or outputs.  


\begin{wrapfigure}{}{0.5\textwidth}
  \begin{center}
    \includegraphics[width=\linewidth]{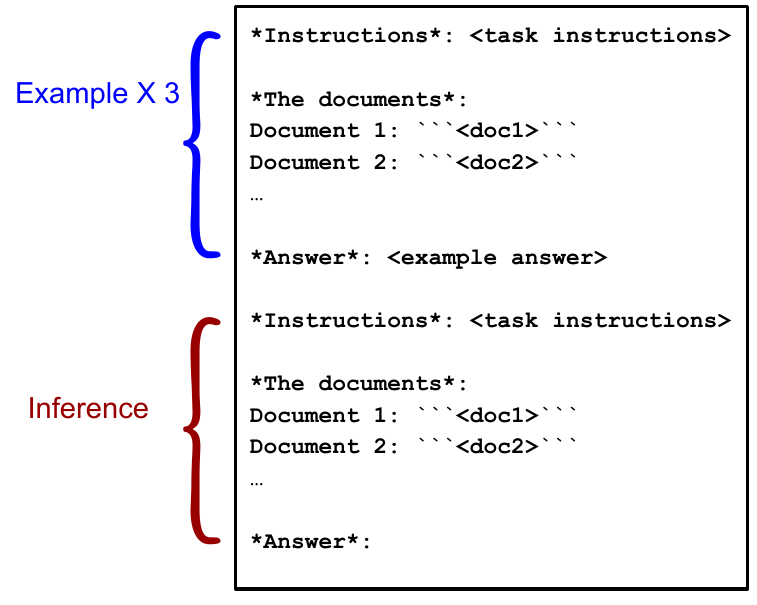}
  \end{center}
  \caption{The \md{} prompt template used for \name{}. Elements in brackets represent template elements filled according to the specific task, dataset, and sample.}
  \label{fig:prompt-template}
\end{wrapfigure}

Concretely, \name{} consists of 6 different datasets, covering different multi-document tasks in various domains, as detailed in Table~\ref{tab:datasets}.
To allow LLM evaluation, we start from a common template for all \md{} tasks, outlined in Figure~\ref{fig:prompt-template}, containing slots for task instruction, documents, and in-context few-shot examples. 

Below we list the set of tasks and corresponding datasets, as well as the required steps we implement in order to include them in the benchmark.

\subsection{Multi-Document Summarization}


\textbf{Task definition.}
Multi-document summarization (MDS) is the task of producing an abridged version of a set of topically related documents, that concisely conveys the salient information in the source texts. Salience is determined by the requirement of the summary consumer, e.g., focusing on the central issues or, alternatively, focusing on a specific request. Summarizing a set of documents requires consolidating information  across the sources, which
introduces challenges such as redundancy, building relations between 
complementing events, and handling conflicting information, 
and accordingly generating a respective output. 
MDS is relevant for a variety of applications
such as summarizing a news event \citep{fabbri-etal-2019-multi}, obtaining information about a desired medical issue \citep{baumel_2016_tdqfs}, or learning about an aspect of a hotel from its reviews~\citep{angelidis_2021_abs_space}.

\textbf{Datasets.}
We include three variants of MDS in \name{}. First, \textit{\mn}~\citep{fabbri-etal-2019-multi} is a prominent generic MDS dataset consisting of news articles from various sources and human-written gold summaries. Each sample includes 2 to 10 source articles, with most samples containing 2-3 documents, and a respective human summary. 
The summaries are abstractive and therefore require synthesizing an output based on information from multiple sources, rather than merely extracting and concatenating sentences from the input documents.
\begin{table}[]
\centering
\caption{Datasets included in the \name{} benchmark. The two right-most columns (number of tokens and number of documents) present the mean values over the entire dataset.}
\label{tab:datasets}
\begin{tabular}{@{}lllrrrr@{}}
\toprule
\multirow{2}{*}{\textbf{Task Family}} &
  \multirow{2}{*}{\textbf{Dataset}} &
  \multirow{2}{*}{\textbf{Domain}} &
  \multirow{2}{*}{\textbf{\# Samples}} &
  \multicolumn{2}{c}{\textbf{\# Toks/Sample}} &
  \multirow{2}{*}{\textbf{\begin{tabular}[c]{@{}c@{}}\# Docs\\ Per Sample\end{tabular}}}\\
                                        &          &            &        & Input & Output &                \\ \midrule
\multirow{3}{*}{Summarization}          & \mn      & News       & 56,100 & 2,349 & 277    & 2.8   \\ 
                                        & \asp     & News       & 1,183  & 8,192 & 102    & 9.9           \\  
                                        & \fuse    & Reviews    & 742    & 704   & 164    & 8.0   \\ \midrule
QA                      & \musique & Wiki  & 44,710 & 2,148 & 17     & 20.0     \\ \midrule
\multirow{2}{*}{Coreference} & \ecb     & News       & 43     & 2,503 & 1,309  & 22.7     \\
                                        & \scico   & Scientific & 521    & 4,207 & 150    & 47.5    \\\bottomrule 
\end{tabular}
\end{table}

Second, \textit{\asp}~\citep{amar-etal-2023-openasp} is a recent dataset for \md{} open aspect-based summarization. In this task, the target summary is expected to focus on a specified aspect within a topic (e.g., a summary focusing on ``launch to orbit'' regarding a document-set on ``Hubble Telescope''). In conventional aspect-based summarization datasets, an aspect can be chosen from a small pre-defined list of domain-thematic aspects (such as ``background'' for any news event).
In contrast, in \asp{} the aspects change with respect to the source documents, consequently requiring a model to process any given aspect that may not have been seen before.
\asp{} was developed upon the DUC~\citep{nist2002DUCWebsite} and \mn{}~\citep{fabbri-etal-2019-multi} datasets in the news domain.

Third, we include \textit{\fuse}~\citep{slobodkin2024multireview}. 
This dataset addresses the task of fusion-in-context, which can be viewed as a step within an MDS pipeline.
Each instance in the \fuse{} dataset consists of 8 reviews of the same business, with spans highlighted within the reviews. The task is then to generate a coherent passage which fuses together all highlighted spans, with minimal additional information which does not originate from  these spans. 
A notable challenge here is that the output passage needs to integrate conflicting opinions, which is commonplace in the subjective reviews. E.g., one customer may commend the cleanliness of a hotel room, while another might criticize it.



\textbf{Task formatting.}
\Md{} summarization readily lends itself to LLM evaluation, as the input primarily consists of raw documents, while the output similarly consists of unstructured natural language. We follow the \md{} prompt template as presented in Figure~\ref{fig:prompt-template}, with dataset-specific instructions. For \asp{} and \fuse{}, we use the instructions provided with the 
respective datasets, while for \mn{}, we use a slightly modified version of the instructions of \asp{}. See the Appendix for full prompts for each of these tasks.



\subsection{Multi-Hop Question Answering}

\textbf{Task definition.} 
Multi-hop question answering requires models to make multiple reasoning steps. Unlike single-hop questions, which can be answered from a single sentence or document, multi-hop questions necessitate processing of data from various sources, e.g., different documents, to arrive at the correct answer.

\textbf{Datasets.}
\name{} includes \textit{\musique}~\citep{trivedi-etal-2022-musique}, which is a dataset of questions based on Wikipedia texts. To answer a question, the model needs to integrate information from 2-4 documents, out of 20 documents provided as its evidence. We use the full version of the dataset, which consists of both answerable and unanswerable questions, where answerability is determined based on weather there is sufficient information in the provided evidence documents to deduce the answer. While the answerable questions test the model's multi-hop reasoning performance, the unanswerable questions both quantify the extent of model hallucinations, and hint at potential cases of memorization.

\textbf{Task formatting.}
We add three prompt elements to adapt our \md{} prompt (Figure~\ref{fig:prompt-template}) to multi-hop QA. First, we add the question after the task instructions and before the documents. Second, we direct the model, via the task instructions and few-shot examples,  to output a JSON format with a dedicated boolean field to indicate whether the model found the question to be answerable from the provided context.



\subsection{Cross-Document Coreference Resolution}
\textbf{Task definition.}
Cross-document coreference resolution aims to identify and link mentions of the same concept (entity or event) across multiple documents. This task is fundamental for applications like multi-document summarization, question answering, and knowledge base construction, where drawing connections across diverse sources is essential for generating accurate and complete outputs. Beyond the challenges of the standard task of within-document coreference resolution, the cross-document task poses its own challenges, notably because documents are authored independently by different authors in a different context. As a result, lexical similarity is not a strong signal, since different expressions can refer to the same concept (e.g., ``label'' and ``response'' in machine learning), whereas similar texts can refer to different concepts (e.g., ``model'' in different scientific fields).

\textbf{Datasets.}
We include two cross-document coreference resolution datasets in \name{}, which vary in domain, number of input documents, and  expected output length. First, we include \textit{\ecb}~\citep{cybulska-vossen-2014-using}, which covers the news domain and is specifically compiled such that the documents discuss similar topics, to increase the challenge introduced via high lexical overlap.

Second, we include \textit{\scico}~\citep{cattan2021scico}, which is a cross-document coreference dataset in the scientific domain, primarily consisting of computer science paper abstracts and some full-text papers about artificial intelligence. Mentions in \scico{} are abstract scientific concepts, which exhibit a high level of lexical diversity (e.g., ``autonomous vehicle'' and ``self driving cars''), and ambiguity (e.g., ``transformers'' in machine learning vs. electrical engineering).

\textbf{Task formatting.}
Coreference resolution typically involves two  tasks: mention detection and coreference linking. 
To simplify the evaluation of LLMs on this task, we follow~\citet{le2023large}, and focus the task only on coreference linking, thus decoupling the evaluation of the aforementioned subtasks~\citep{cattan-etal-2021-realistic}. 
We provide the oracle mentions as part of the input by marking the mention spans within the input text with unique identifiers.
We format the output as a list of clusters, where each cluster is a list of coreferring mention identifiers. See a full example in the Appendix.

\section{A Stochastic Approach to LLM Evaluation}
\label{sec:stochastic}

Popular benchmarks typically evaluate all models against \emph{a single prompt template} per task For example, each dataset in the popular MMLU benchmark \citep{hendrycks2021measuring} has a a single instruction explaining the task, the expected output format,  a fixed set of possible answers per sample, ordered and formatted in one particular manner. 

A recent line of work has noticed that current LLMs are highly sensitive to seemingly arbitrary prompt design choices. These include semantically-equivalent paraphrases of the task instructions (e.g., replacing ``term'' with ``word'')~\cite{mizrahi2024state}, minor prompt formatting variations (e.g., replacing spaces with new lines, or adding commas)~\cite{sclar2023quantifying,voronov2024mind}, the order of given in-context few-shot examples~\cite{lu-etal-2022-fantastically}, the instances chosen for evaluation~\cite{Perlitz2023EfficientB}, or the order of the documents in multi-document tasks~\cite{levy2024task,10.1162/tacl_a_00638}. These works show that variations along any of these axes lead to vast differences in both absolute and relative model performance, thus calling into question many of the observations made based on prevalent single-prompt benchmarks~\cite{liang2023holistic,srivastava2023beyond,suzgun2022challenging}.

To achieve a more meaningful evaluation, we propose to account for the sensitivity of LLMs as a core component within benchmarks. In particular, we follow~\citet{machery_2019}, who suggests that experimental design should differentiate between two types of factors. First, some observed variables are claimed to be \emph{causally related} to a target experimental variable. For example, the choice of the architecture, pretraining data, instruction tuning, and other modelling design choices are expected to have causal effects on the performance of models in multi-document tasks. Second, \emph{unrelated variables} are presumed not to have a causal relation with the observed variable. In the case of LLMs, such variables include those arbitrary variables discussed above, for which there is not a clear methodology for preferring one choice over another (e.g., different prompt formats), and they do not seem to make the task easier or harder for a human annotator. To achieve a meaningful, 
\emph{replicable} experiment, \citet{machery_2019} argues that variables which are considered unrelated to the outcome should be resampled at random in repeated experiments to assess the experiment's reliability.

We implement this approach by designing \name{} as a stochastic \emph{instance generator},
rather than being a fixed static collection of instances. We lay out a set of configuration variables for a prompt, along with a space of possible value options for each variable. We then construct $k$ evaluation sets 
by resampling different assignments, uniformly at random. This resampling enables us to examine model performance via a host of statistical measures, including its mean, standard deviation, maximum or minimum performance, and compare them against other models on the same dataset instances and configurations. 

\textbf{Sampled variables.}
We identify five unrelated variables, based on recent observations on the sensitivity of LLMs to prompt variations~\cite{mizrahi2024state,sclar2023quantifying,voronov2024mind}. For each of these, we also list the possible space of options from which we sample uniformly at random to create different dataset instances. 

First, we sample from a space of possible \emph{instruction paraphrases}. Since this space is hard to enumerate, we employ the instruction paraphrasing approach proposed by~\citet{mizrahi2024state} to obtain a list of 20 paraphrases for each task instruction. 
Second, a specific \emph{order of $n$ presented documents} is sampled at random from all of the possible $n!$ permutations. 
Third, $k$ few-shot examples are sampled from a pool of 5 possible examples which we assemble from each dataset's training portion. Fourth, we choose a random order of the few-shot examples out of the possible $k!$ permutations. Finally, the fifth randomly sampled variable is the set of dataset instances on which to evaluate, which are sampled i.i.d without replacement from each dataset's test or validation portions. 

\section{Evaluation}
\label{sec:results}

Below we describe the LLMs we evaluate on \name{} (\ref{sec:models}), the experimental setup for evaluation (\ref{sec:setup}), followed by the results and analyses over the different tasks (\ref{sec:res}).

\subsection{Models}\label{sec:models}

We evaluate 7 instruction-tuned LLMs on \name, from several model families, 
including \mistral~\citep{jiang2023mistral}, \mixtral~\citep{jiang2024mixtral}, \llama~\citep{llama3modelcard} and \gemma~\citep{team2024gemma}, as outlined in Table~\ref{tab:models}, and Table~\ref{tab:appendix-models} in the Appendix. Due to
computational constraints we run a quantized (4 bit) version of \mixtral-8x22B.

\begin{wraptable}{r}{0.38\textwidth}
\centering
\caption{Models evaluated on \name.}
\label{tab:models}
\begin{tabular}{@{}lrr@{}}
\toprule
\textbf{Name} & \textbf{\# Params} & \textbf{\begin{tabular}[c]{@{}l@{}}Context\\ Window\end{tabular}} \\ \midrule
\multirow{2}{*}{\mixtral} & 8x7B  & 32k \\
                          & 8x22B & 32k \\\midrule
\mistral                  & 7B    & 8k  \\\midrule
\multirow{2}{*}{\llama}   & 8B    & 8k  \\
                          & 70B   & 8k  \\\midrule
\multirow{2}{*}{\gemma}   & 2B    & 8k  \\
                          & 7B    & 8k  \\ \bottomrule
\end{tabular}
\end{wraptable}

We opt not to include closed, API-based, models (e.g., OpenAI models) for two main reasons. First, these models have been shown to wrap the input prompts with meta-prompts, and perform further manipulations~\citep{rao2023tricking}, which can confound the results of our stochastic prompt choice. Second, evaluating these models on \name{} is expensive to run over long inputs. For example, evaluating GPT4 on \name{} with our strategy would cost roughly 100USD for a single dataset within the benchmark. We make \name{} freely available and welcome future work to report more models on it.

\subsection{Experimental Setup}\label{sec:setup}
\name{} is configured by four hyper-parameters: (1) we use 0.8 as the decoding temperature for all evaluated LLMs, as used in previous LLMs evaluation~\cite{chen2021evaluating}; (2) following recent work on LLM sensitivity to prompt formatting~\cite{sclar2023quantifying}, we report results over 10 \emph{i.i.d resamplings} of the arbitrary choices discussed in the previous section; (3) we choose to include $k = 3$ \emph{few-shot examples} with each prompt, note that the specific examples and their order are resampled with each evaluation run; and (4) we sample a maximum of 100 \emph{input instances} from each dataset, where here too the specific instances are resampled every run. 


\textbf{Metrics.} For \mn{} and \asp{} we  compute the geometric mean of ROUGE-1, ROUGE-2 and ROUGE-L scores, as done in SCROLLS~\cite{shaham-etal-2022-scrolls}. For \fuse{}, we use the F1 score, which is the harmonic mean between the faithfulness and the coverage scores, as established by \citet{slobodkin2024multireview}. For \scico{} and \ecb{}, we evaluate each topic independently, following ~\citet{cattan-etal-2021-realistic}, and report the traditional CoNLL-F1 score for coreference resolution. 
For \musique{}, we measure the F1 overlap between gold and predicted outputs, as suggested in the dataset's paper.

\begin{table}[]
\centering
\caption{Model ranking for the different datasets and tasks in \name{}. Following~\citet{dror-etal-2019-deep}, the column marked `$\downarrow$' denotes the pairwise stochastic dominance between two consecutively-ranked models. The column marked `\#' is $Rank_i(m)$, which is the rank of model $m$ in task $i$. $^{*}$due to computational constraints we run a quantized (4 bit) version of \mixtral-8x22B.}
\label{tab:ranking}
\resizebox{\textwidth}{!}{%
\begin{tabular}{@{}llllllllllll@{}}
\toprule
\multicolumn{12}{c}{\textbf{Summarization}} \\
\multicolumn{4}{c}{\textbf{\mn}} &
  \multicolumn{4}{c}{\textbf{\asp}} &
  \multicolumn{4}{c}{\textbf{\fuse}} \\ \midrule
\multicolumn{1}{c}{$\downarrow$} &
  \multicolumn{1}{c}{\#} &
  \multicolumn{1}{c}{Model} &
  \multicolumn{1}{c|}{Mean  $\pm$  std} &
  \multicolumn{1}{c}{$\downarrow$} &
  \multicolumn{1}{c}{\#} &
  \multicolumn{1}{c}{Model} &
  \multicolumn{1}{c|}{Mean  $\pm$  std} &
  \multicolumn{1}{c}{$\downarrow$} &
  \multicolumn{1}{c}{\#} &
  \multicolumn{1}{c}{Model} &
  \multicolumn{1}{c}{Mean  $\pm$  std} \\ \midrule
 &
  1 &
  \mixtral-8x7B &
  \multicolumn{1}{l|}{21.4 $\pm$ 0.7} &
   &
  1 &
  \mistral-7B &
  \multicolumn{1}{l|}{11.8 $\pm$ 0.6} &
   &
  1 &
  \mistral-7B &
  77.5 $\pm$ 1.7 \\
1.0 &
   &
   &
  \multicolumn{1}{l|}{} &
  1.0 &
   &
   &
  \multicolumn{1}{l|}{} &
  0.82 &
   &
   &
   \\
 &
  2 &
  \llama-70B &
  \multicolumn{1}{l|}{20.9 $\pm$ 0.5} &
   &
  2 &
  \mixtral-8x7B &
  \multicolumn{1}{l|}{11.5 $\pm$ 0.5} &
   &
  2 &
  \llama-70B &
  76.8 $\pm$ 1.6 \\
1.0 &
   &
   &
  \multicolumn{1}{l|}{} &
  1.0 &
   &
   &
  \multicolumn{1}{l|}{} &
  1.0 &
   &
   &
   \\
 &
  3 &
  \mixtral-8x22B$^*$ &
  \multicolumn{1}{l|}{20.3 $\pm$ 0.4} &
   &
  3 &
  \mixtral-8x22B$^*$ &
  \multicolumn{1}{l|}{11.1 $\pm$ 0.3} &
   &
  3 &
  \llama-8B &
  75.9 $\pm$ 1.3 \\
0.99 &
   &
   &
  \multicolumn{1}{l|}{} &
  0.97 &
   &
   &
  \multicolumn{1}{l|}{} &
  1.0 &
   &
   &
   \\
 &
  4 &
  \mistral-7B &
  \multicolumn{1}{l|}{20.1 $\pm$ 0.6} &
   &
  4 &
  \llama-70B &
  \multicolumn{1}{l|}{9.0 $\pm$ 2.0} &
   &
  4 &
  \mixtral-8x7B &
  65.7 $\pm$ 5.5 \\
0.97 &
   &
   &
  \multicolumn{1}{l|}{} &
  1.0 &
   &
   &
  \multicolumn{1}{l|}{} &
  0.66 &
   &
   &
   \\
 &
  5 &
  \llama-8B &
  \multicolumn{1}{l|}{19.5 $\pm$ 0.9} &
   &
  5 &
  \llama-8B &
  \multicolumn{1}{l|}{5.6 $\pm$ 0.5} &
   &
  5 &
  \mixtral-8x22B$^*$ &
  65.0 $\pm$ 2.4 \\
1.0 &
   &
   &
  \multicolumn{1}{l|}{} &
  1.0 &
   &
   &
  \multicolumn{1}{l|}{} &
  1.0 &
   &
   &
   \\
 &
  6 &
  \gemma-2B &
  \multicolumn{1}{l|}{8.7 $\pm$ 1.0} &
   &
  6 &
  \gemma-2B &
  \multicolumn{1}{l|}{2.9 $\pm$ 0.4} &
   &
  6 &
  \gemma-2B &
  59.6 $\pm$ 4.6 \\
1.0 &
   &
   &
  \multicolumn{1}{l|}{} &
  1.0 &
   &
   &
  \multicolumn{1}{l|}{} &
  1.0 &
   &
   &
   \\
 &
  7 &
  \gemma-7B &
  \multicolumn{1}{l|}{2.4 $\pm$ 0.2} &
   &
  7 &
  \gemma-7B &
  \multicolumn{1}{l|}{2.5 $\pm$ 0.3} &
   &
  7 &
  \gemma-7B &
  38.5 $\pm$ 3.0 \\ \midrule
 &
   &
   &
   &
   &
   &
   &
   &
   &
   &
   &
   \\ \midrule
\multicolumn{4}{c|}{\textbf{Question Answering}} &
  \multicolumn{8}{c}{\textbf{Coreference Resolution}} \\
\multicolumn{4}{c|}{\textbf{\musique}} &
  \multicolumn{4}{c}{\textbf{\scico}} &
  \multicolumn{4}{c}{\textbf{\ecb}} \\ \midrule
\multicolumn{1}{c}{$\downarrow$} &
  \multicolumn{1}{c}{\#} &
  \multicolumn{1}{c}{Model} &
  \multicolumn{1}{c|}{Mean  $\pm$  std} &
  \multicolumn{1}{c}{$\downarrow$} &
  \multicolumn{1}{c}{\#} &
  \multicolumn{1}{c}{Model} &
  \multicolumn{1}{c|}{Mean  $\pm$  std} &
  \multicolumn{1}{c}{$\downarrow$} &
  \multicolumn{1}{c}{\#} &
  \multicolumn{1}{c}{Model} &
  \multicolumn{1}{c}{Mean  $\pm$  std} \\ \midrule
 &
  1 &
  \llama-70B &
  \multicolumn{1}{l|}{57.3 $\pm$ 1.3} &
   &
  1 &
  \mistral-7B &
  \multicolumn{1}{l|}{31.1 $\pm$ 1.3} &
   &
  1 &
  \llama-70B &
  22.3 $\pm$ 3.4 \\
1.0 &
   &
   &
  \multicolumn{1}{l|}{} &
  1.0 &
   &
   &
  \multicolumn{1}{l|}{} &
  0.67 &
   &
   &
   \\
 &
  2 &
  \llama-8B &
  \multicolumn{1}{l|}{49.4 $\pm$ 1.9} &
   &
  2 &
  \llama-70B &
  \multicolumn{1}{l|}{24.3 $\pm$ 1.7} &
   &
  2 &
  \llama-8B &
  21.9 $\pm$ 1.9 \\
1.0 &
   &
   &
  \multicolumn{1}{l|}{} &
  0.72 &
   &
   &
  \multicolumn{1}{l|}{} &
  1.0 &
   &
   &
   \\
 &
  3 &
  \mixtral-8x22B$^*$ &
  \multicolumn{1}{l|}{45.8 $\pm$ 3.8} &
   &
  3 &
  \llama-8B &
  \multicolumn{1}{l|}{24.1 $\pm$ 0.8} &
   &
  3 &
  \mistral-7B &
  20.1 $\pm$ 0.9 \\
1.0 &
   &
   &
  \multicolumn{1}{l|}{} &
  1.0 &
   &
   &
  \multicolumn{1}{l|}{} &
  1.0 &
   &
   &
   \\
 &
  4 &
  \gemma-2B &
  \multicolumn{1}{l|}{17.2 $\pm$ 2.8} &
   &
  4 &
  \mixtral-8x22B$^*$ &
  \multicolumn{1}{l|}{21.5 $\pm$ 2.7} &
   &
  4 &
  \mixtral-8x7B &
  14.1 $\pm$ 3.8 \\
1.0 &
   &
   &
  \multicolumn{1}{l|}{} &
  1.0 &
   &
   &
  \multicolumn{1}{l|}{} &
  1.0 &
   &
   &
   \\
 &
  5 &
  \mistral-7B &
  \multicolumn{1}{l|}{11.5 $\pm$ 4.8} &
   &
  5 &
  \mixtral-8x7B &
  \multicolumn{1}{l|}{17.8 $\pm$ 3.4} &
   &
  5 &
  \mixtral-8x22B$^*$ &
  12.3 $\pm$ 2.5 \\
1.0 &
   &
   &
  \multicolumn{1}{l|}{} &
  1.0 &
   &
   &
  \multicolumn{1}{l|}{} &
  1.0 &
   &
   &
   \\
 &
  6 &
  \mixtral-8x7B &
  \multicolumn{1}{l|}{5.6 $\pm$ 1.2} &
   &
  6 &
  \gemma-2B &
  \multicolumn{1}{l|}{3.4 $\pm$ 0.8} &
   &
  6 &
  \gemma-2B &
  4.3 $\pm$ 1.8 \\
1.0 &
   &
   &
  \multicolumn{1}{l|}{} &
  0.93 &
   &
   &
  \multicolumn{1}{l|}{} &
   &
   &
   &
   \\
 &
  7 &
  \gemma-7B &
  \multicolumn{1}{l|}{0.8 $\pm$ 0.4} &
   &
  7 &
  \gemma-7B &
  \multicolumn{1}{l|}{3.3 $\pm$ 0.8} &
   &
  7 &
  \gemma-7B &
  NA \\ \bottomrule
\end{tabular}%
}
\end{table}

\begin{figure}[htb]
    \centering 
\begin{subfigure}{.5\textwidth}
  \includegraphics[width=\linewidth]{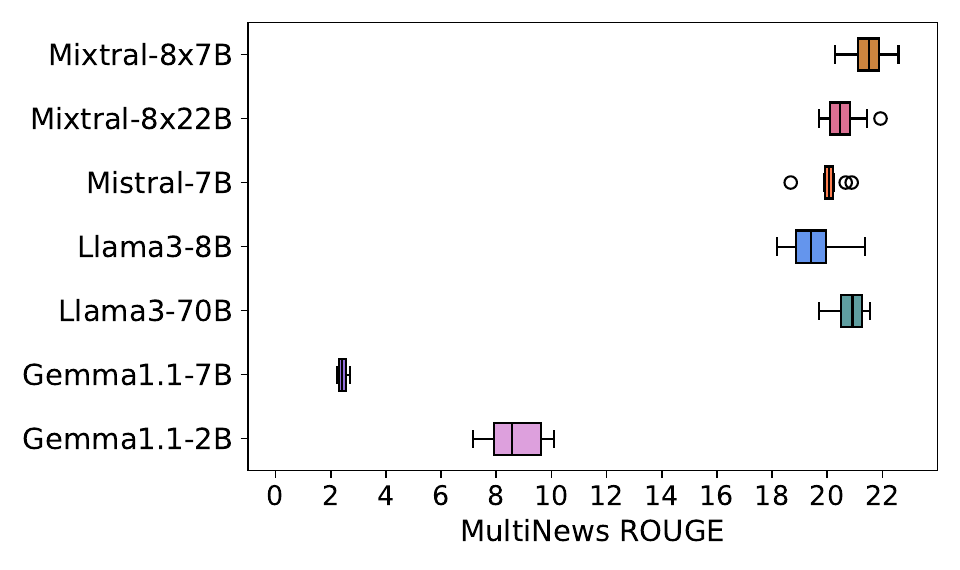}
\end{subfigure}\hfil 
\begin{subfigure}{.5\textwidth}
  \includegraphics[width=\linewidth]{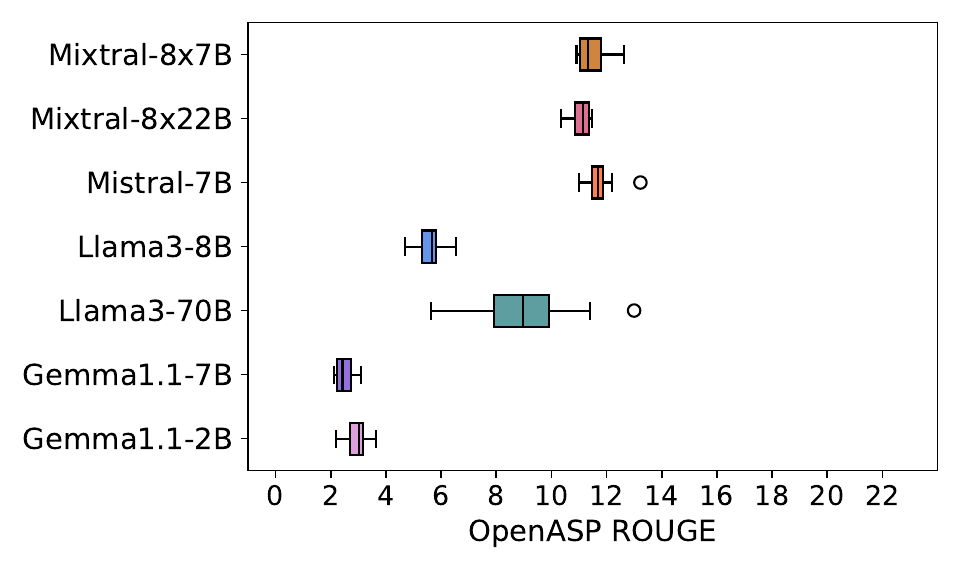}
\end{subfigure}\hfil 
\begin{subfigure}{.5\textwidth}
  \includegraphics[width=\linewidth]{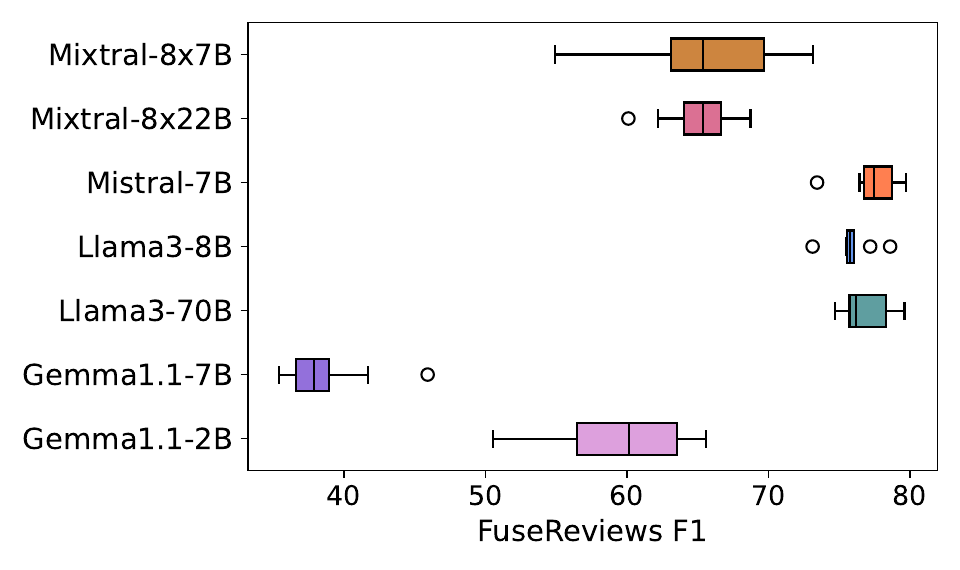}
\end{subfigure}\hfil 
\begin{subfigure}{.5\textwidth}
  \includegraphics[width=\linewidth]{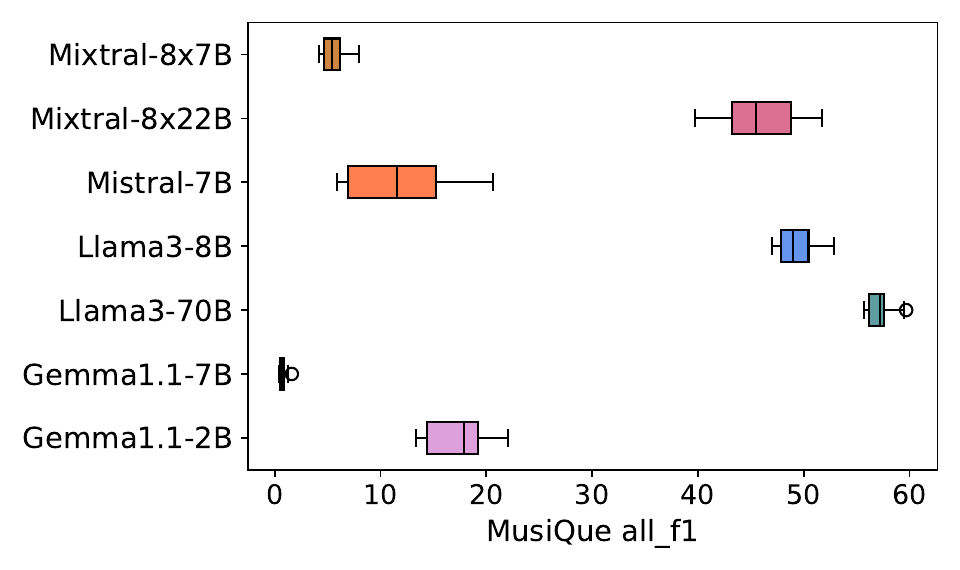}
\end{subfigure}\hfil 
\begin{subfigure}{.5\textwidth}
  \includegraphics[width=\linewidth]{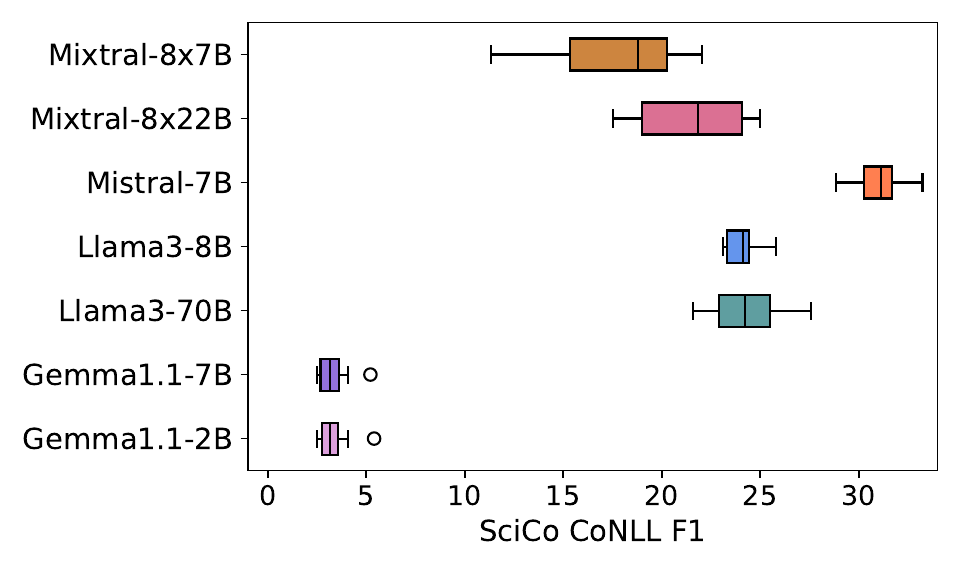}
\end{subfigure}\hfil 
\begin{subfigure}{.5\textwidth}
  \includegraphics[width=\linewidth]{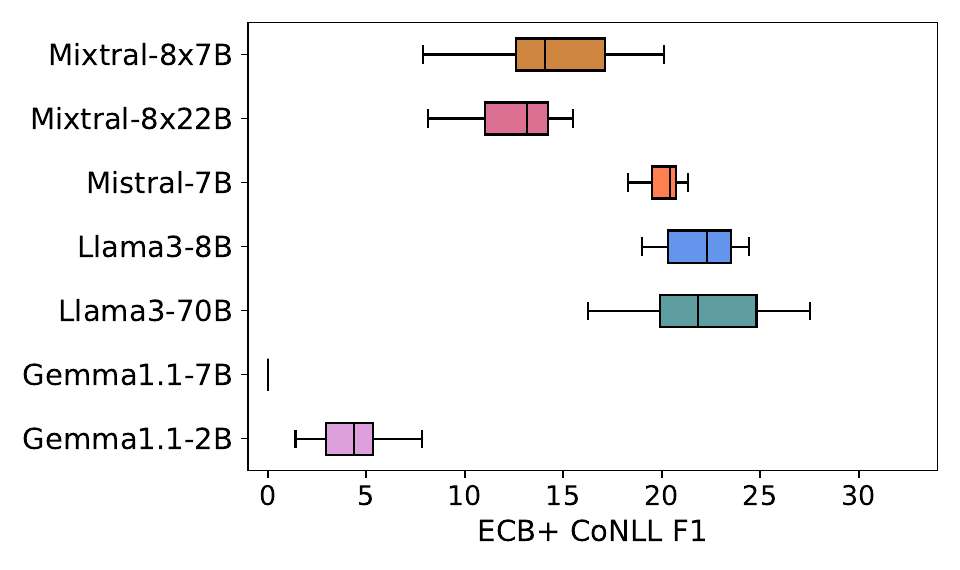}
\end{subfigure}\hfil 
\caption{Result distributions across different tasks. Each box-plot shows the distribution of 10 different resamplings for each model, on each task.}
\label{fig:boxplots-main}
\end{figure}

\subsection{Results}\label{sec:res}
\begin{figure}
  \centering 
\begin{subfigure}{.333\textwidth}
  \includegraphics[width=\linewidth]{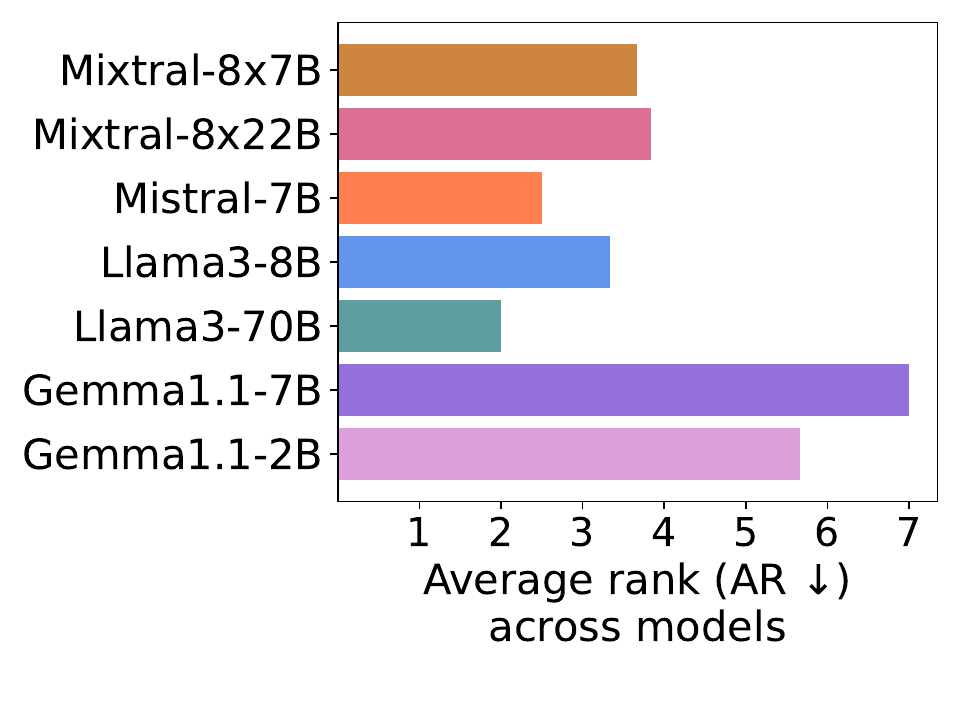}
  \label{fig:ranking}
\end{subfigure}\hfil 
\begin{subfigure}{.333\textwidth}
  \includegraphics[width=\linewidth]{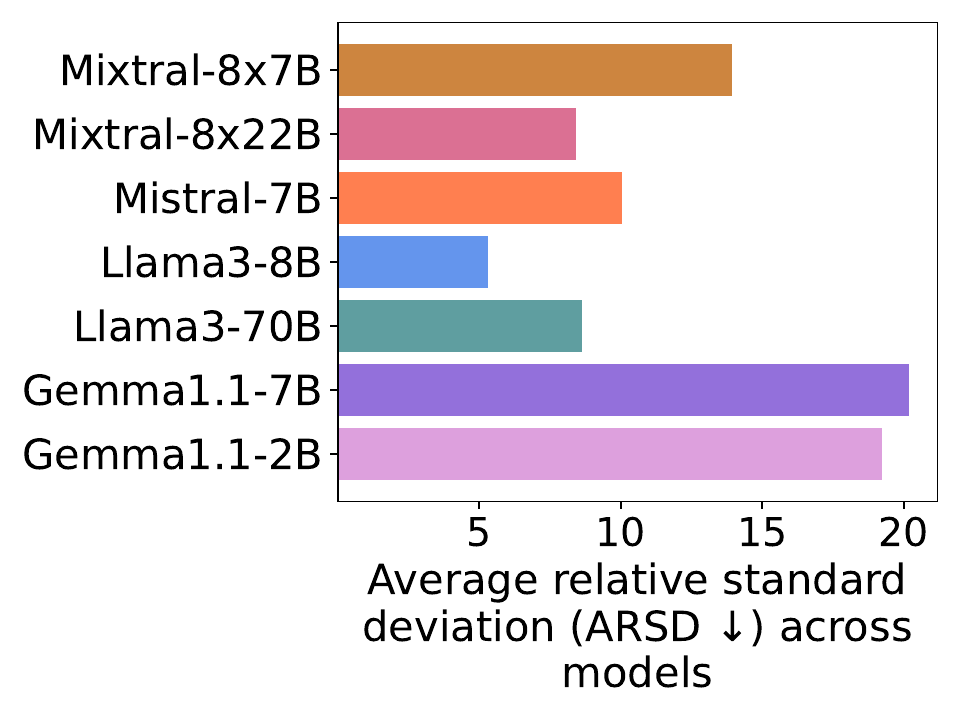}
  \label{fig:std}
\end{subfigure}\hfil 
\begin{subfigure}{.333\textwidth}
  \includegraphics[width=\linewidth]{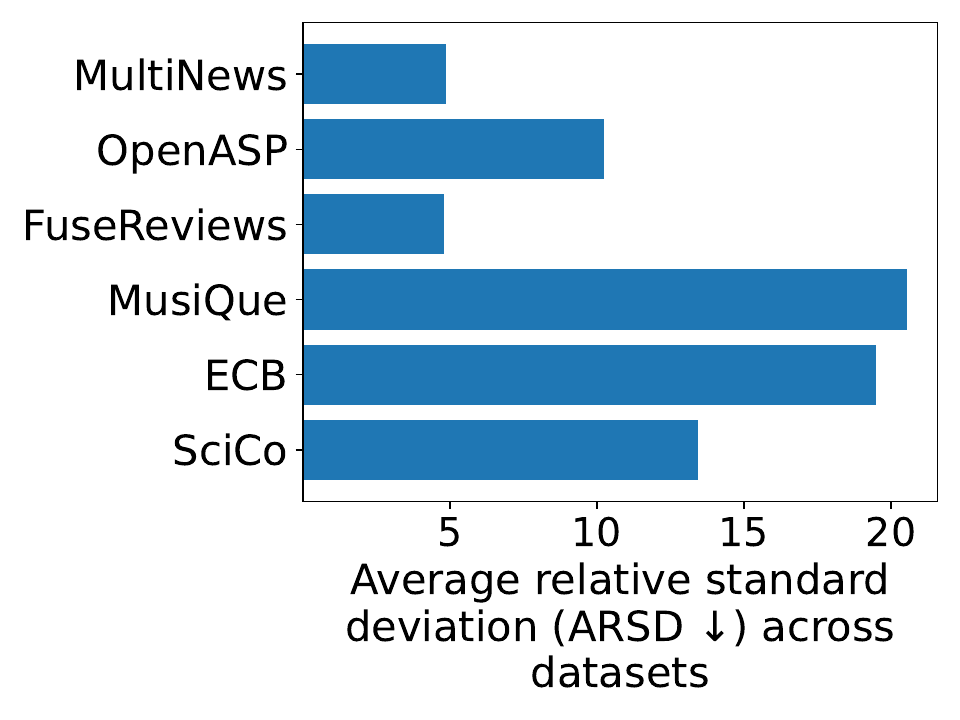}
  \label{fig:std-by-task}
\end{subfigure}
\caption{The averaged rank ($AR$) and relative standard deviation ($ARSD$) based on the mean score of all executions, averaged over all datasets. 
Lower $ARSD$ indicates higher robustness to prompt variations. Lower $AR$ indicates more accurate performance relative to other models.}
  \label{fig:acg-rank}
\end{figure}

Our experiments yield predictions over  resamplings of seemingly arbitrary design choices, such as instruction phrasing or few-shot ordering, enabling a statistical comparison between models over these choices. 
In Table~\ref{tab:ranking} we present model rankings $Rank_i(m)$ for each model $m$ and dataset $i$, based on mean performance across 10 resamplings, and Figure~\ref{fig:boxplots-main} shows the distribution of model results for each dataset.
In addition, in Figure~\ref{fig:acg-rank} we show aggregated performance across datasets, computed in terms of average rank ($AR$) and averaged relative standard deviation ($ARSD$; \cite{everitt2010cambridge}), which are defined as follows:

\[
\begin{array}{cc}
  ARSD(m) \eqdef \frac{1}{d}\sum\limits_{i = 1}^d\frac{\sigma_i(m)}{\mu_i(m)}; & AR(m) \eqdef  \frac{1}{d}\sum\limits_{i = 1}^d Rank_i(m) \\
 
\end{array}
\]

Where $d = 6$ is the number of datasets in \name{}, and for a given LLM $m$, $\mu_i(m)$ denotes the mean empirical performance of $m$ on dataset $i$,  $\sigma_i(m)$ is the empirical standard deviation for the same dataset $i$, and recall that $Rank_i(m)$ is $m$'s ranking on dataset $i$, relative to all other models. All of these metrics are computed across $r = 10$ resamples.

Below we derive few interesting observations from the results of LLMs on \name{}.


\textbf{The \md{} tasks included in \name{} leave significant room for improvement in LLMs.} The results in Table~\ref{tab:ranking}, suggests that  \name{} remains far from saturation and could be a valuable tool for tracking the advancement of LLMs in handling \md{} tasks in the future. Follow up work may compare these numbers to human performance on these tasks to assess an upper limit for performance.

\textbf{Model performance varies widely in different resamplings of arbitrary choices.}
Consistent with previous findings~\cite{sclar2023quantifying,mizrahi2024state}, Figure~\ref{fig:boxplots-main} shows that LLM performance varies with different formatting and phrasing of instructions.
This variability is further illustrated in Figure~\ref{fig:acg-rank}, where we aggregate model standard deviations across all tasks.
To quantify this variation, we employ a variation of the stochastic dominance metric suggested by \citet{dror-etal-2019-deep}, which captures the stochastic nature of the results for more accurate model comparisons. 
We propose that leaderboards should also include the probability that one model outperforms another model, as shown in the ‘$\downarrow$’ column in Table~\ref{tab:ranking}. While some tasks show clear superiority between consecutively ranked models, such as in \musique{}, others, like \fuse{}, exhibit more nuanced differences. 



\textbf{The challenge in \md{} tasks is not explained by the context length.}
We observe that  performance does not correlate with input length. This suggests that the challenge in multi-document tasks is the consolidation of information from across multiple documents, and not necessarily the increasing length of input. Supporting evidence is presented in Figure~\ref{fig:input_len_for_appendix} in the Appendix. 

\textbf{Summarization datasets are less sensitive to arbitrary design choices than other tasks.}
In Figure~\ref{fig:acg-rank} we compute the averaged relative standard deviation \emph{across datasets} by averaging model performance on each of them. We observe that summarization-related datasets (\asp, \mn{}, and \fuse{}) tend to be more robust to variations in prompt. This can be a result of the unstructured nature of its output, while other tasks, like coreference-resolution and question answering, require a structured format, such that small changes in the prediction may have higher effect on the scoring than in the natural output format of summarization.

\textbf{LLMs vary in their ability to follow an expected output format.} 
We find a correlation between a model's ability to predict the expected output format for \musique, \scico, and \ecb, to its overall performance, where $p$ is the Pearson coefficient. These findings imply that future work that focuses on improving LLMs' performance in tasks requiring specific output formatting and post-processing, should first focus on models' ability to infer the expected format from the input prompt. 


\textbf{Larger model size does not guarantee better performance.} 
Even though \llama-70B has ten times the number of parameters compared to \mistral-7B,  Table~\ref{tab:ranking} shows that \mistral-7B outperforms \llama-70B in \asp, \fuse, and \scico. Similarly, in terms of sensitivity to prompt variations, some smaller models provide a more robust response, as depicted in Figure~\ref{fig:acg-rank} where \llama-8B has lower $ARSD$ compared to \llama-70B.

\section{Related Work}
\label{sec:related}


Scrolls and ZeroScrolls~\citep{shaham-etal-2022-scrolls, shaham-etal-2023-zeroscrolls} are benchmarks for tasks requiring long context inputs. \Md{} tasks pose a unique challenge that does not appear in long context. 
MMLU~\cite{hendrycks2021measuring} is a benchmark that became a standard for evaluating LLMs. As opposed to \name{}, all tasks are formed in a multiple-choice QA format. 
HELM~\cite{liang2023holistic} benchmark includes a robustness score to minor variations in input, e.g., single word misspelling, while \name{} offers extensive paraphrasing and various design choices. Moreover, HELM reports accuracy on a single static variation, while in \name{}, each instance is obtained via a resampling and results are reported with statistical measures across these choices.

\section{Conclusions}
We introduced \name{}, a stochastic benchmark for \md{} tasks. \name{} allows to evaluate LLMs specifically on \md{} tasks, while taking into account arbitrary design choices, such as task instructions paraphrasing, what few-shots examples to introduce within the prompt and in which order, the order of documents for the \md{} input, and which instances for each task will be given to the model. We find that the \md{} datasets included in \name{} leave room for improvement for LLMs, stating \name{} as a relevant and useful benchmark for LLMs. In addition, we introduce a new methodological approach to address the sensitivity of LLMs to arbitrary choices in their prompt, that can be adapted to other benchmarks beyond \name{}. 

\ack

We thank Sergey Feldman, Doug Downey and Rotem Dror for their insightful feedback and consultations. We thank and appreciate the cloud GCP credits granted by Google Research for this project.

\bibliographystyle{abbrvnat}
\bibliography{costum}


\appendix
\section{Appendix}

\subsection{Limitations}\label{sec:limitaitions}
\name{} includes existing datasets, some of which are relatively older (e.g., \mn{} from 2019). It is possible that the older datasets have leaked into training sets of today's SOTA LLMs, including the LLMs we experiment on in this paper. However, due to \name{}'s stochastic nature, and the random design choices we sample in each run, it is reasonable to assume that models were not exposed to all possible variations as we test them. If they eventually do, alternative paraphrases or output formats can be substituted in order to maintain \name's relevance. Furthermore, we observe that some advanced LLMs that we tested are still lagging significantly behind others.

Regarding societal impacts, we chose to include in \name{} datasets that are reported to not contain harmful or offensive content. However, since \name{} includes vast amounts of texts that we have little control over, it is possible that such content has unintentionally snuck into the datasets.

\subsection{Reproducing Results on \name{}}\label{sec:reproducing}
To reproduce results on our benchmark, our code base produces a random seed associated with each benchmark run. This seed can be used to reproduce evaluations and get comparable results for new LLMs on identically-sampled choices. Below we provide the JSON configuration file (Listing~\ref{config-json}), used to run the experiments reported in this paper.

\begin{listing}
\begin{minted}[frame=single,
               framesep=3mm,
               linenos=false,
               xleftmargin=21pt,
               tabsize=4]{js}
{
  "out_dir": "<PATH_TO_OUT_DIR>",
  "run_name": "<NAME>",
  "random_seed": 42,
  "num_different_runs": 10,
  "num_demonstrations": 3,
  "max_num_samples": 100,
  "datasets": [
    {
      "name": "ECB",
      "task": "coreference resolution",
      "path": "<LOCAL_PATH_TO_ECB>",
      "split_name": "test_events"
    },
    {
      "name": "FuseReviews",
      "task": "Fusion in-context",
      "split_name": "validation"
    },
    {
      "name": "SciCo",
      "task": "coreference resolution",
      "split_name": "test"
    },
    {
      "name": "MultiNews",
      "task": "summarization",
      "split_name": "test"
    },
    {
      "name": "MusiQue",
      "task": "question answering",
      "path": "<LOCAL_PATH_TO_MusiQue>",
      "split_name": "val"
    },
    {
      "name": "OpenASP",
      "task": "summarization",
      "path": "<LOCAL_PATH_TO_OpenASP>",
      "split_name": "test"
    }
  ]
}
\end{minted}
\caption{configuration JSON used for the resampling of \name{}.} 
\label{config-json}
\end{listing}

\subsection{Technical Details}\label{sec:appendix-setup}
In Table~\ref{tab:appendix-models} we provide the specific model versions and hardware used for running the reported experiments in this paper.
\begin{table}[ht!]
\centering
\caption{Technical details regarding the models evaluated on \name. \mixtral-8x22 is used in its 4-bit quantized version.}
\label{tab:appendix-models}
\begin{tabular}{@{}llll@{}}
\toprule
\textbf{Name}             & \textbf{\# Params} & \textbf{HuggingFace model}                     & \textbf{Hardware} \\ \midrule
\multirow{2}{*}{\llama}   & 8B                & meta-llama/Meta-Llama-3-8B-Instruct   & 1xA100-40GB       \\
                          & 70B               & meta-llama/Meta-Llama-3-70B-Instruct  & 8xA100-40GB       \\ \midrule
\mistral                  & 7B                & mistralai/Mistral-7B-Instruct-v0.2    & 1xA100-40GB       \\ \midrule
\multirow{2}{*}{\mixtral} & 8x7B              & mistralai/Mixtral-8x7B-Instruct-v0.1  & 4xA100-40GB       \\
                          & 8x22B             & mistralai/Mixtral-8x22B-Instruct-v0.1 & 8xA100-40GB       \\ \midrule
\multirow{2}{*}{\gemma}   & 2B                & google/gemma-1.1-2b-it                & 1xA100-40GB       \\
                          & 7B                & google/gemma-1.1-7b-it                & 1xA100-40GB       \\ \bottomrule
\end{tabular}
\end{table}

\subsection{Licensing of the Included Datasets}\label{sec:licensing}
\asp{} has an MIT license, for \ecb{} it is CC BY 3.0. \musique{}, \scico{} and \fuse{} has an Apache 2.0 license. From \mn{} license within their github repo: ``...The Dataset is intended for non-commercial research and educational purposes only, and is made available free of charge without extending any license or other intellectual property rights.''\footnote{\url{https://github.com/Alex-Fabbri/Multi-News?tab=License-1-ov-file}}

\subsection{Analysis of Performance w.r.t. Input Length}\label{sec:appendix-additional-results}

\begin{figure}[htb]
    \centering 
    \begin{subfigure}{.33\textwidth}
  \includegraphics[width=\linewidth]{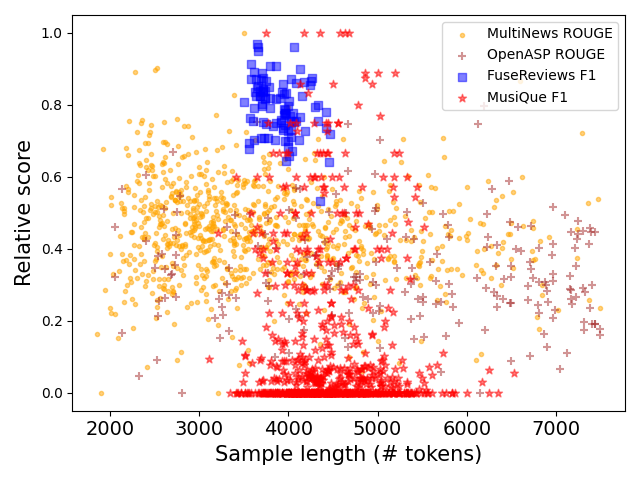}
  \caption{\mistral-7B}
  \label{fig:len-mistral-7b}
\end{subfigure}\hfil 
\begin{subfigure}{.33\textwidth}
  \includegraphics[width=\linewidth]{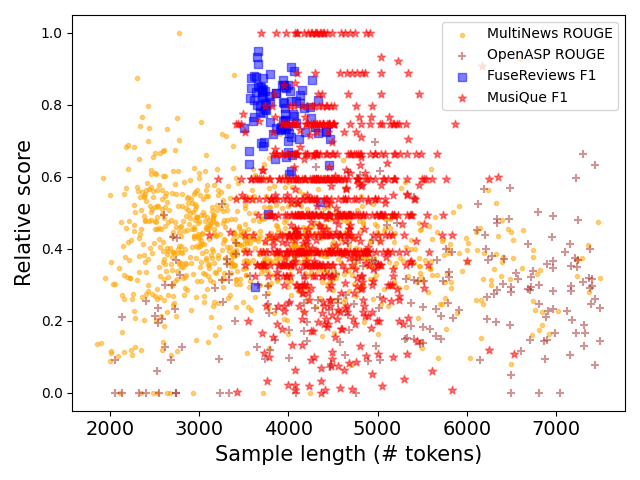}
  \caption{\llama-8B}
  \label{fig:len-llama-8b}
\end{subfigure}\hfil 
\begin{subfigure}{.33\textwidth}
  \includegraphics[width=\linewidth]{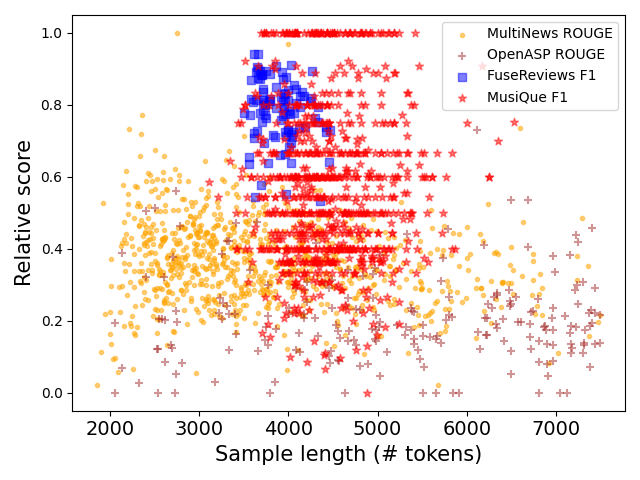}
  \caption{\llama-70B}
  \label{fig:len-llama-70b}
\end{subfigure}\hfil 
\vspace{0.4cm}
\begin{subfigure}{.33\textwidth}
  \includegraphics[width=\linewidth]{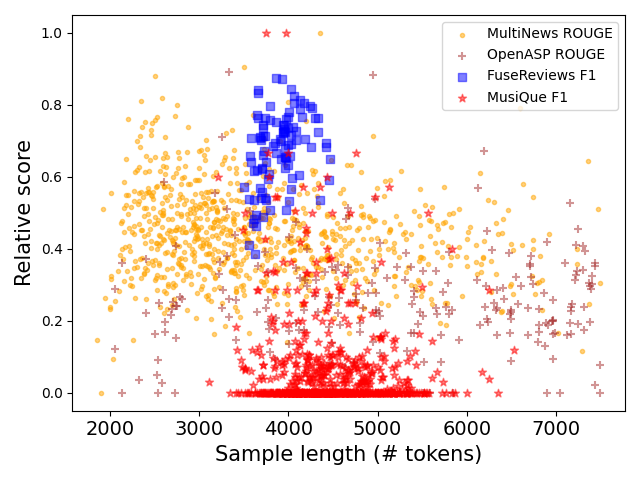}
  \caption{\mixtral-8x7B}
  \label{fig:len-mixtral-8x7B}
\end{subfigure}\hfil 
\begin{subfigure}{.33\textwidth}
  \includegraphics[width=\linewidth]{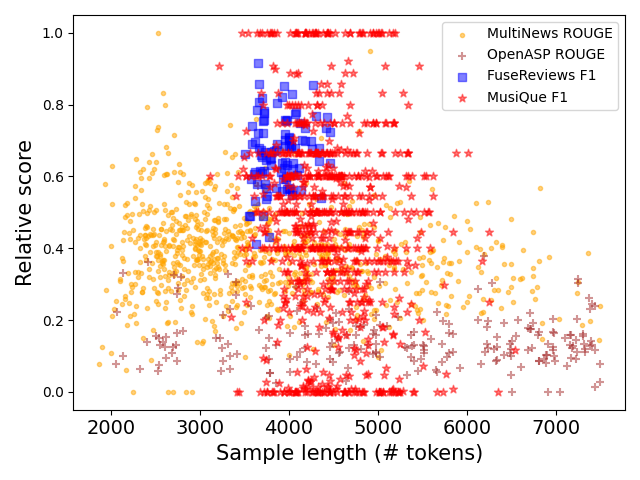}
  \caption{\mixtral-8x22B}
  \label{fig:len-mixtral-8x22B}
\end{subfigure}\hfil 
\vspace{0.4cm}
\begin{subfigure}{.33\textwidth}
  \includegraphics[width=\linewidth]{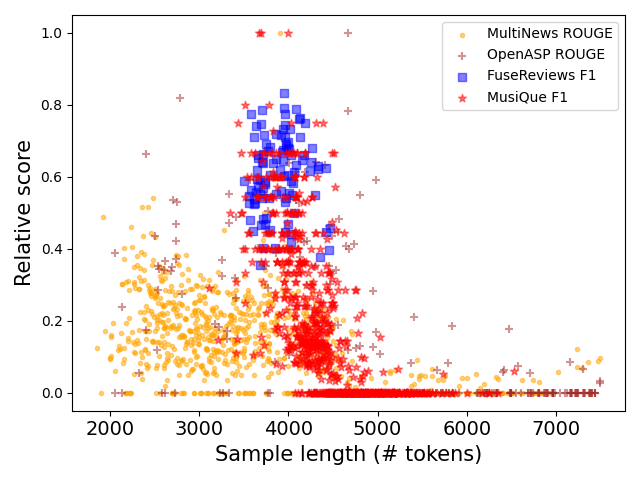}
  \caption{\gemma-2B}
  \label{fig:len-gemma-2B}
\end{subfigure}\hfil 
\begin{subfigure}{.33\textwidth}
  \includegraphics[width=\linewidth]{images/input_length/polyfit_Gemma1.1-2B.png}
  \caption{\gemma-7B}
  \label{fig:len-gemma-7B}
\end{subfigure}\hfil 
\caption{Performance as a function of input length.}
\label{fig:input_len_for_appendix}
\end{figure}

In Figure~\ref{fig:input_len_for_appendix} we examine the effect of each instance's length, on model performance. The x-axis is the length of input (in \# tokens), and the y-axis is the relative performance score for each task (i.e., normalized between min and max performance for that task). 
As seen in the figure, no correlation is seen between input length and output scores, supporting the claim that the challenge within \name{} lies in handling multiple documents in the input, and not necessarily in its input length.

\subsection{Analysis of Performance w.r.t. Output Format}
To analyze tasks requiring specific output formats, such as QA or coreference resolution, we investigate the correlation between the model's ability to predict the correct format and its overall performance. We calculate the Pearson correlation coefficient to quantify this relationship, and the results are as follows:
$$
\rho_{MusiQue}=0.99;\:\: \rho_{ECB+}=0.86;\:\: \rho_{SciCo}=0.93
$$
These high coefficients support our claim that future work on such tasks should first focus on models’ ability to infer the expected format from the input prompt.

\newpage
\section{Instruction Paraphrases}

Below is the full set of instruction paraphrases we sample from, for each task in \name{}.

\subsection{\mn}

1. "In this task, you are presented with multiple news articles about related topics. Your job is to generate an extractive summary that integrates information from the provided articles. Your summary should be short and concise, that includes content only from the provided articles, avoiding any external data sources."

2. "Please provide a brief, extractive summary by synthesizing only the key points from the articles provided. Focus on the main arguments and conclusions without incorporating any information from outside these texts. Keep your summary concise and directly related to the content of the documents."

3. "Generate a concise extractive summary using only the information from the provided articles. Your summary should distill the most essential information, capturing the core insights without adding any external content. Aim for brevity and clarity in your summarization."

4. "Please sift through the provided articles and distill their essence into a sharp, concise summary. Focus solely on the facts and key points within these texts, avoiding any embellishment or reference to external information. Your summary should read like a bullet-point list of the most critical insights."

5. "You are presented with multiple news articles about related topics. Summarize the contents in a way that captures the key information in a narrative form, but strictly using the details mentioned in the provided documents. Keep it engaging yet brief."

6. "Imagine you're preparing a brief for a decision-maker who has limited time. Summarize the provided documents by extracting only the most essential information. Present this in a clear, straightforward manner, focusing on the key facts and figures, and ensure that all content is directly sourced from the articles without external references."

7. "Using only the details from the articles I've given you, craft a summary that distills the most important information. Avoid any interpretations or external data, and keep your summary short and direct. Emphasize the main arguments, data points, and conclusions from the documents."

8. "Operate as an information synthesizer: Draw the essence from multiple articles, focusing solely on the information contained within them. Your summary should be a tight, focused digest of the articles, free from any influence of external data."

9. "Scan through the provided articles and compile a summary that highlights only the most significant facts and figures, ensuring the exclusion of all external references. Aim for clarity and brevity."

10. "Operate as an academic summarizer: Imagine you are creating a summary for an academic review. Extract and emphasize the most pertinent information, ensuring your summary remains true to the original texts and free of external content."

11. "Condense the provided information into a compact summary that emphasizes the main points and crucial data from the documents. Exclude any external information to maintain the integrity of the sources."

12. "From the provided articles, pull out the core messages and data points. Shape these into a brief, clear summary that directly reflects the content of the documents without any external additions."

13. "Compile a concise summary from the news articles given, focusing only on the information contained within. Your summary should integrate the main points without adding any outside information."

14. "Create a succinct extractive summary by focusing exclusively on the details provided in the articles. Avoid using any external sources and ensure the summary remains clear and to the point."

15. "Produce a brief summary that distills the essential facts from the provided articles. Keep your summary strictly to the content presented in the documents, avoiding external influences."

16. "Develop a concise extractive summary using only the information from the articles provided. Emphasize the main points and conclusions while avoiding the inclusion of any external data."

17. "Prepare a short, integrated summary by synthesizing key points from the given news articles. Ensure that no external content is included and that the summary is clear and direct."

18. "Your task is to distill the primary information from the provided articles into a concise summary. Make sure to exclude any external sources and focus strictly on the given texts."

19. "Summarize the provided articles by extracting only the key information and conclusions. Your summary should be brief and must not incorporate any external data."

20. "Generate a clear and brief extractive summary using just the information from the provided articles. Focus on distilling the essential points and data without referencing external content."

\subsection{\asp}

1. "In this task you are required to generate an aspect-based summary of a set of documents related the same topic. Please write a short, concise aspect-based summary, only summarize content from the above documents, avoiding any external data sources."

2. "Your goal is to create a short, concise aspect-based summary of the given documents. Summarize the key points accurately, using only the information from these documents and excluding any external sources."

3. "Produce a brief, aspect-based summary of the collection of documents on the same topic. Ensure your summary is concise and derived only from the provided documents, avoiding any external data sources."

4. "Your task is to generate a detailed yet concise aspect-based summary from a collection of documents that focus on the same topic. Begin by thoroughly examining each document to understand the main aspects and themes. Then, synthesize this information into a coherent summary that highlights the significant points. Make sure your summary is short and derived exclusively from the content of the provided documents, without incorporating any external data."

5. "Given a set of documents related to a specific topic, generate a short, concise aspect-based summary. Ensure that the summary is based solely on the content of the documents provided"

6. "You will receive several documents on the same topic. Your task is to write a brief aspect-based summary, using only the information from the provided documents and excluding any external sources."

7. "You are tasked with generating an aspect-based summary of several documents. Summarize the content briefly and accurately, using only the information from the documents give"

8. "In this task, you are required to create an aspect-based summary of a set of documents all related to the same topic. Carefully read through each document and identify the key aspects discussed. Summarize these aspects in a concise manner, ensuring that your summary captures the essential points. It is crucial to base your summary solely on the provided documents, avoiding any external information or references. "

9. "You are tasked with producing an aspect-based summary for a series of documents related to the same topic. Start by analyzing each document to identify the critical aspects covered. Your goal is to condense this information into a clear and concise summary, ensuring that you accurately represent the main points. The summary should be brief and entirely based on the provided documents, with no inclusion of external sources or data."

10. "Generate a concise aspect-based summary of the given documents. Focus on summarizing the content based solely on the information from these documents, avoiding any external sources."

11. "Create a concise aspect-based summary for the provided set of documents. Focus on the main aspects and themes discussed in these documents, ensuring that your summary is based entirely on the content of the provided documents and excludes any external sources."

12. "Produce a short and precise aspect-based summary of the given documents. Identify the key aspects discussed in these documents and synthesize a concise summary based solely on the provided content."

13. "You will receive a collection of documents focused on the same topic. Your task is to create an aspect-based summary that highlights the key aspects discussed in these documents. Ensure your summary is brief and does not include any external information."

14. "You are provided with multiple documents related to a single topic. Your task is to generate an aspect-based summary that captures the main aspects discussed in these documents. Ensure your summary is concise and solely based on the provided texts."

15. "You are tasked with generating an aspect-based summary of several documents on the same topic. Carefully review each document, identify the main aspects, and write a brief summary that captures these aspects using only the provided documents."

16. "Your role is to create an educational summary for students using a collection of documents on the same topic. Focus on the main aspects that would help students understand the core concepts discussed in the documents. Write a short, clear aspect-based summary, relying exclusively on the provided texts."

17. "Imagine you are preparing a briefing for a busy executive who needs to understand the key aspects of several documents quickly. Summarize the most important points from these documents in a concise manner, ensuring your aspect-based summary is derived entirely from the content of the provided documents and avoids any external information."

18. "As an advanced AI tasked with summarizing documents, your goal is to generate an aspect-based summary. Think of yourself as a summarization expert, extracting the most critical aspects from the documents provided. Craft a concise summary that highlights these key aspects, ensuring it is based solely on the given documents."

19. "Imagine you are a journalist tasked with writing a summary article based on a series of documents related to a single topic. Identify the key aspects discussed in these documents and compose a brief, coherent summary that encapsulates the main points without introducing any external information."

20. "Your task is to act as a knowledge distiller, creating a concise aspect-based summary from a series of documents on the same topic. Focus on identifying and summarizing the critical aspects discussed in these documents, ensuring your summary is brief and based exclusively on the provided content."

21. "You are an AI assistant tasked with providing a summary for a set of documents related to a specific topic. Focus on the key aspects and themes discussed in these documents. Create a summary that captures these aspects in a concise manner, ensuring that your summary is based solely on the provided documents and excludes any external information."

\subsection{\fuse}

1. "In this task, you are presented with eight reviews, where some parts are \"highlighted\" (namely, there are {HS} and {HE} tokens before and after each such span). Your job is to generate a summary that covers all and only the \"highlighted\" spans.",
            
 2. "You are presented with eight reviews, each containing highlighted sections marked by {HS} and {HE} tokens. Your task is to carefully read through the reviews and generate a comprehensive summary that accurately captures all the content within the highlighted spans.",

 3. "Examine the eight reviews provided, noting the highlighted sections indicated by the {HS} and {HE} tokens. Construct a summary that encapsulates the key points and details found within these highlighted spans, ensuring a complete and concise overview.",
            
  4. "Read through the eight reviews and identify the highlighted content demarcated by {HS} and {HE}. Your task is to create a thorough summary that includes all the essential points and insights from the highlighted sections.",
            
5. "Review the eight provided reviews and focus on the content between the {HS} and {HE} markers. Produce a summary that effectively captures the main ideas and sentiments expressed within these highlighted spans.",
            
6. "In this task, read the eight reviews and pay special attention to the highlighted sections marked with {HS} and {HE}. Generate a summary that encompasses the insights and viewpoints found in these highlighted portions.",
            
7. "You will be presented with eight reviews that contain highlighted spans marked by {HS} and {HE}. Your goal is to produce a summary that covers all the important aspects and main points from these highlighted sections.",
            
            8. "Analyze the eight reviews and concentrate on the highlighted sections marked with {HS} and {HE}. Create a summary that includes all the relevant points and perspectives found within these highlighted spans.",
            
            9. "From the eight reviews, focus on the highlighted spans marked with {HS} and {HE}. Your task is to generate a summary that encapsulates all the critical details and information found within these highlighted sections.",
            
            10. "Read through the eight reviews and identify the text marked by {HS} and {HE}. Create a summary that accurately represents all the key points and sentiments found within these highlighted portions.",
           
            11. "You are provided with eight reviews that contain highlighted sections marked with {HS} and {HE}. Generate a summary that includes all the main points, observations, and opinions from these highlighted spans.",

            12. "Examine the eight reviews and focus on the text enclosed by the {HS} and {HE} markers. Produce a summary that presents a comprehensive overview of the important points and insights found within these highlighted areas.",
            
            13. "You will read through eight reviews that contain highlighted sections marked by {HS} and {HE}. Your goal is to create a summary that captures the key points and themes found within these highlighted spans.",
            
            14. "Read the eight reviews carefully and focus on the text within the {HS} and {HE} tokens. Your task is to create a detailed summary that covers all the important aspects found in these highlighted sections.",
            
            15. "Examine the eight provided reviews and focus on the sections marked by {HS} and {HE}. Generate a summary that effectively represents the main points and ideas from these highlighted spans.",
            
            16. "You have eight reviews that contain highlighted spans marked by {HS} and {HE}. Create a summary that includes all the key details, observations, and insights found within these highlighted areas.",
            
            17. "Read the eight reviews and concentrate on the highlighted sections marked by {HS} and {HE}. Your task is to create a summary that encapsulates the most important aspects found in these highlighted spans.",
            
            18. "Examine the eight reviews and focus on the highlighted sections marked with {HS} and {HE}. Produce a summary that effectively captures all the critical details and main ideas from these highlighted spans.",
            
            19. "From the eight reviews, focus on the highlighted sections marked by {HS} and {HE}. Your task is to create a summary that encompasses all the essential points and insights found within these highlighted spans.",
            
            20. "Read through the eight reviews and concentrate on the text bracketed by {HS} and {HE}. Generate a summary that captures the key ideas and observations found within these highlighted portions.",
            "comments": "Task focuses on capturing key ideas and observations in the summary."

\subsection{\musique}

1. "In this task, you are presented with question, and 20 documents that covers the answer to that question. Deduce your answer solely from the provided documents, avoiding any external data sources. Keep the answer short and concise, leave behind any irrelevant details. If the documents don't have the answer, set 'is\_answerable' to false in the output dictionary. If they do, set 'is\_answerable' to true and put the answer in 'answer\_content'."

2. "Answer the given multi-hop question, based only on information from the 20 documents provided below. Task: multi hop question answering from a multi document input. Output: The answer should be deduced solely from the text from the above documents, avoiding any external data sources. Give a short, to-the-point response, excluding irrelevant details.  Use 'is\_answerable': false in the output dictionary if the answer isn't in the documents. If it is, set 'is\_answerable': true and include the answer in 'answer\_content'."

3. "Input: A multi-hop question and a collection of multi-document texts. Task: Understand the question and decompose it into its sub questions. Answer the decomposed questions from the provided multiple documents. Come up with the answer to the original input question. Output: A short and concise answer to the question, leave behind any irrelevant details. Also, the answer should be deduced solely from the text from the above documents, avoiding any external data sources. Indicate 'is\_answerable': false in the output if the answer isn't found in the documents. If found, use 'is\_answerable': true and place the answer in 'answer\_content'."

4. "Your task is to read text from multi-document input, and answer a multi-hop question solely from the information that is in the documents. Input: A multi-hop question and a collection of multi-document texts. Task: Answer the question provided from the collection of multi-documents. Output: A short and concise answer to the question, leave behind any irrelevant details. Also, the answer should be deduced solely from the text from the above documents, avoiding any external data sources. Set 'is\_answerable': false in the dictionary if the documents don't contain the answer. If they do, set 'is\_answerable': true and write the answer in 'answer\_content'."

5. "Input: A complex, multi-hop question alongside a collection of related documents. Task: Carefully analyze the question to identify its underlying sub-questions. Utilize the provided documents to systematically address each sub-question. Synthesize the answers to these sub-questions to formulate a comprehensive response to the original query. Output: Deliver a succinct and precise answer to the initial question. Ensure that the response is derived entirely from the provided documents, disregarding any information from external sources. If the documents lack the answer, use 'is\_answerable': false in the output dictionary. If the answer is there, use 'is\_answerable': true and include it in 'answer\_content'."

6. "Your task to to deliver a succinct and precise answer to the given question, out of a given multiple document texts. Ensure that the response is derived entirely from the provided documents, disregarding any information from external sources. Mark 'is\_answerable' as false if the answer isn't in the documents. If it is, mark 'is\_answerable' as true and provide the answer in 'answer\_content'."

7. "Imagine you're a detective with a pile of evidence before you. Your case? To solve the puzzle posed by a given question. Use only the clues from these documents to build your case and crack the question wide open. No outside help—just you and the documents. Set 'is\_answerable': false in the output if the documents don't have the answer. If they do, set 'is\_answerable': true and include the answer in 'answer\_content'."

8. "You are an advanced AI capable of processing and analyzing text from multiple documents related to a specific topic. Your expertise involves extracting only the essential information to construct a clear and concise answer to complex questions. Below are multiple documents on a unified theme, each separated by triple backticks. Your task is to answer the provided question. Focus your analysis to derive the answer strictly from the content provided within these documents, ensuring no external information is used. Keep the reply short and clear, leaving out irrelevant parts. Indicate 'is\_answerable' as false if the answer can't be found in the documents. If found, use 'is\_answerable' as true and add it to 'answer\_content'."

9. "You are an AI model specialized in multi-hop question answering. Your function is to meticulously read through multiple documents and synthesize information to answer questions that require connecting data across different texts. Enclosed are several documents, each marked by triple backticks. The challenge is to answer a multi-layered question. Ensure that your response is entirely based on the documents, with no external data influencing the outcome. Deliver a short, clear response, excluding unnecessary information. If no answer is found in the documents, use 'is\_answerable': false. If the answer is found, use 'is\_answerable': true and provide it in 'answer\_content'."

10. "You function as a thematic synthesizer, equipped with the ability to discern thematic patterns and extract critical insights from multiple documents. Your role is to fuse this information into a coherent response that addresses a specified multi-layered question. Provided here are multiple documents, each themed around a central topic and individually separated by triple backticks within different user messages. Your challenge is to distill these texts into a precise answer to the given question. Your response must be formed exclusively from the contents of these documents, avoiding any external influences. If the answer isn't in the documents, set 'is\_answerable': false in the output. If it is, set 'is\_answerable': true and put the answer in 'answer\_content'."

11. "You task is to answer a multi-hop question given multiple documents. You are an expert AI designed for contextual inquiry, capable of navigating complex information landscapes to pinpoint accurate answers. Your core skill lies in evaluating and interpreting interconnected data from diverse documents to address specific questions. Your objective is to analyze the given documents comprehensively and provide a focused response to the given question, relying solely on the data within these texts. Ensure your response is brief and to the point, avoiding extra information. Use 'is\_answerable': false if the documents don't have the answer. If they do, set 'is\_answerable': true and include it in 'answer\_content'."

12. "As a Data Integration Specialist AI, your primary function is to seamlessly integrate data from various texts, focusing on extracting relevant information to answer intricate questions effectively. You excel at forming concise, fact-based answers from a compilation of detailed documents. Here are several documents, each distinctively marked by triple backticks. Engage with these texts to form a coherent and succinct answer to the given question. Your analysis should strictly use the provided documents as the sole information source. Mark 'is\_answerable' as false if you can't find the answer in the documents. If the answer is available, set 'is\_answerable' as true and provide it in 'answer\_content'."

13. "Welcome to your research challenge! Today, you are provided with multiple documents, each enclosed by triple backticks. Your mission is to delve into these texts and uncover the answer to the given question. Remember, your investigation should draw solely from these documents. Happy researching! Set 'is\_answerable' to false if the answer isn't found in the documents. If it is, set it to true and place the answer in 'answer\_content'."

14. "Your task is to synthesize information from multiple documents and answer the provided multi-hop question. Carefully read through the documents to extract essential information, connect relevant data points, and arrive at a concise, accurate answer. Base your response solely on the information found in the documents, avoiding any external sources. If the answer isn't in the provided documents, set 'is\_answerable' to false. If it is, set 'is\_answerable' to true and include it in 'answer\_content'."

15. "Think of yourself as an evidence collector. Your goal is to navigate through multiple documents and analyze them to deduce the answer to the given question. Examine each document thoroughly, and form a concise and accurate response using information only from the provided texts. Indicate 'is\_answerable' as false if the documents lack the answer. If they contain it, set 'is\_answerable' as true and provide the answer in 'answer\_content'."

16. "Your task is to conduct a logical analysis of multiple documents to answer the multi-hop question. Carefully review each document, drawing connections and inferences to synthesize an accurate and concise response. Avoid any external information; your answer should be based solely on the provided texts. Use 'is\_answerable': false if the answer can't be found in the documents. If it can, use 'is\_answerable': true and provide it in 'answer\_content'."

17. "You are tasked with solving a multi-layered question using the information provided in multiple documents. Examine each text carefully to find relevant information and connections. Form a concise and well-reasoned response using only the provided documents, without referencing external sources. If no answer is found in the documents, set 'is\_answerable' to false. If an answer is found, set it to true and include it in 'answer\_content'."

18. "As an inquiry specialist, your task is to navigate through multiple documents to answer the given question. Scrutinize each text for relevant information, make connections between them, and formulate a concise answer based solely on the provided documents. Set 'is\_answerable' as false in the output dictionary if the documents don't have the answer. If they do, set 'is\_answerable' as true and provide it in 'answer\_content'."

19. "Your task is to synthesize data from multiple documents to answer the multi-hop question. Read through the texts carefully, identify key details and themes, and combine them to form a concise response. Ensure your answer is derived solely from the provided documents, avoiding any external sources. Mark 'is\_answerable' as false if the answer isn't in the documents. If it is, mark 'is\_answerable' as true and include it in 'answer\_content'."

20. "Approach the question as a detective piecing together evidence from multiple documents. Read each text thoroughly, identify patterns and relationships, and deduce a clear, concise answer. Base your response solely on the provided documents, avoiding any external data. Set 'is\_answerable' to false if the answer isn't found in the documents. If it is, set it to true and place the answer in 'answer\_content'."

21. "Your task is to navigate through multiple documents and synthesize information to answer a multi-hop question. Examine each document carefully to identify key details and connections. Form a clear, concise answer using only the provided texts, avoiding any external sources. Use 'is\_answerable': false in the output dictionary if the answer isn't in the documents. If it is, set 'is\_answerable': true and include the answer in 'answer\_content'.
"

\subsection{\scico}

1. "In this task you are presented with multiple documents that are related to the same topic. In each document, each mention of an entity or event is enclosed with square brackets, followed by the mention's unique id. That is: '[some mention](id)'. Your task is to identify and cluster together mentions that are coreferring with each other, i.e., mentions that mark the same entity or event. output the list of clusters, where each cluster is a list of ids of coreferring mentions. For example, if you think that mentions (1) and (2) refer to the same entity, and mentions (3) and (4) refer to the same entity, but these two entities are different, you should output: [[1, 2], [3, 4]]."

2. "Given a set of documents related to the same topic, each containing entities or events marked with unique identifiers in the format '[some mention](id)', your task is to analyze these documents and identify mentions that refer to the same entity or event. Once identified, group these mentions into clusters, where each cluster contains the ids of mentions that are coreferential. Output the clusters in the following format: a list of lists, where each inner list contains the ids of coreferring mentions. For instance, if mentions (1) and (2) are coreferential, and mentions (3) and (4) are coreferential, but refer to different entities, the output should be: [[1, 2], [3, 4]]. Please process the input and produce the correct clusters based on coreference."

3. "Your task is to process a set of documents, each discussing a similar topic and containing marked mentions of entities or events in the format '[some mention](id)'. Identify mentions that refer to the same entity or event and group these ids into clusters. Each cluster should consist of ids that belong to coreferential mentions. Output these clusters in a nested list format, for example, if mentions tagged as (1) and (2) are coreferential, and (3) and (4) are as well, but separate from the first group, your output should look like: [[1, 2], [3, 4]]."

4. "Examine several documents linked by a common theme, noting that each mention of an entity or event within these documents is denoted by '[some mention](id)'. Your objective is to find and group ids of mentions that correspond to the same entity or event into clusters. The desired output format is a list of lists, where each sublist contains ids of mentions that are coreferential. For example, an appropriate output for coreferential mentions (1) and (2), and another pair (3) and (4), would be [[1, 2], [3, 4]]."

5. "Within multiple documents on a similar subject, each entity or event mention is bracketed and tagged with an id in the format '[some mention](id)'. Your role is to ascertain which mentions refer to the same entity or event, grouping these ids into clusters. Each cluster will be a list containing the ids of mentions that are coreferential. For instance, if you determine that (1) and (2), as well as (3) and (4), refer to distinct entities, you should produce the output as: [[1, 2], [3, 4]]."

6. "Analyze a series of documents that discuss the same topic where entities and events are noted as '[some mention](id)'. Group mentions that refer to the same entity or event into clusters by their ids. Each cluster should be presented as a list of these ids, such as [[1, 2], [3, 4]], indicating coreference between the mentions within each list."

7. "Review a collection of documents with related topics, identifying and clustering ids of entity and event mentions, denoted as '[some mention](id)'. Mentions that are coreferential should be grouped together. Format your output as a series of lists, each containing ids of mentions that reference the same entity or event, e.g., [[1, 2], [3, 4]]"

8. "Given multiple documents related by a common theme, each containing marked mentions of entities or events as '[some mention](id)', identify which mentions refer to the same entity or event and cluster these mentions' ids together. Your objective is to find and group ids of mentions that correspond to the same entity or event into clusters. The desired output format is a list of lists, where each sublist contains ids of mentions that are coreferential. For example, an appropriate output for coreferential mentions (1) and (2), and another pair (3) and (4), would be [[1, 2], [3, 4]]."

9. "Given a collection of documents on the same topic, identify mentions of entities or events marked as '[some mention](id)'. Your task is to group ids of mentions that refer to the same entity or event into clusters. Output these clusters as lists of ids, each representing a set of coreferential mentions. For example, if you find that mentions (1) and (2) refer to the same entity, and (3) and (4) refer to another, output should be: [[1, 2], [3, 4]]."

10. "Analyze a set of documents that discuss a unified topic, where each entity or event is uniquely identified by '[some mention](id)'. Group together the ids of mentions that are coreferential. Present your findings as a list of lists, where each sublist contains ids of mentions referring to the same entity or event, e.g., [[1, 2], [3, 4]]. Ensure accuracy in identifying and clustering these mentions."

11. "Review documents related by a common theme, each containing marked mentions of entities or events in the format '[some mention](id)'. Determine which mentions refer to the same entity or event and cluster their ids accordingly. Output these clusters as nested lists, where each list contains ids of coreferential mentions, such as [[1, 2], [3, 4]]. This task requires precise identification and clustering based on coreference."

12. "Your challenge is to navigate through multiple documents related by a common topic, identifying mentions of entities or events enclosed as '[some mention](id)'. Group these mentions by ids when they refer to the same entity or event. Present your clusters as lists of ids that show coreference, for instance, [[1, 2], [3, 4]]. This task requires keen attention to detail and accuracy in identifying connections"

13. "Act as a data detective and delve into several documents, each marked by similar themes and containing entities or events tagged as '[some mention](id)'. Your mission is to uncover which mentions are coreferential and cluster their ids accordingly. Your findings should be reported as lists of ids grouped by coreference, like [[1, 2], [3, 4]]."

14. "Task: Analyze a collection of thematic documents to identify and cluster mentions of entities or events, formatted as '[some mention](id)', that refer to the same concept. Your output should clearly display clusters of ids representing coreferential mentions, such as [[1, 2], [3, 4]], demonstrating your ability to discern and link related information."

15. "In this detailed analysis task, you are provided with multiple documents, each discussing a similar theme. Within these documents, mentions of entities or events are specifically highlighted and labeled with unique identifiers in the format 'some mention](id)'. Your primary objective is to meticulously identify which of these mentions refer to the same real-world entity or event, and then systematically group together their corresponding ids into clusters. Each cluster should exclusively contain ids of mentions that are coreferential. For clarity in your output, format your results into a list of these clusters, where each sublist represents a distinct group of coreferential mentions. For instance, if your analysis concludes that mentions (1) and (2) are about the same entity, and mentions (3) and (4) about another distinct entity, your output should be formatted as: [[1, 2], [3, 4]]. This task requires precise attention to detail and analytical rigor to ensure accuracy in the clustering process."

16. "Examine several documents on a unified topic, noting each mention of an entity or event as '[some mention](id)'. Group mentions by ids that refer to the same entity or event. Output these clusters in a format of nested lists, with each list containing ids of coreferential mentions, like [[1, 2], [3, 4]]."

17. "In this comprehensive task, you are faced with an array of documents tied by a common thematic element. Each document features various mentions of entities or events, each enclosed within square brackets and followed by a unique identification number in the format '[some mention](id)'. It is your responsibility to sift through these mentions, discerning which ones are references to the same entity or event across different texts. Upon identifying these coreferential mentions, you are to organize and cluster their ids into coherent groups. The final output should be a structured list of these clusters, with each cluster formatted as a list containing ids of mentions that you have determined to be about the same entity or event. For example, if you deduce that mentions (1) and (2) discuss one entity, and mentions (3) and (4) another, your result should be presented as: [[1, 2], [3, 4]]. This task demands a high level of accuracy and a methodical approach to ensure that each cluster is correctly assembled based on coreference analysis."

18. "Within a group of related documents, each mention of an entity or event is uniquely identified in the format '[some mention](id)'. Your objective is to examine these documents and cluster the ids of mentions that are coreferential, i.e., refer to the same entity or event. The output should be a list of lists, where each sublist contains the ids of mentions that are grouped together based on coreference. For example, if mentions (1) and (2) are about the same entity, and (3) and (4) are about another, then the output should look like: [[1, 2], [3, 4]]."

19. "Engage in a task where you need to sift through several documents that deal with similar topics, noting that each entity or event mention is tagged in the format '[some mention](id)'. Identify and group mentions that refer to the same entity or event into clusters by their ids. Produce an output that consists of lists, where each list is a cluster of ids representing coreferential mentions. For instance, if (1) and (2) are deemed coreferential, as are (3) and (4), then your output should be structured as: [[1, 2], [3, 4]]."

20. "Your task involves processing a set of documents on a common subject, where each document includes mentions of entities or events marked by '[some mention](id)'. Your role is to determine which mentions refer to the same entity or event and cluster their ids together. The expected output format is a series of nested lists, with each list containing ids of coreferential mentions. For example, if mentions (1) and (2) are about the same entity, and (3) and (4) are about another, organize your output as: [[1, 2], [3, 4]]."

\subsection{\ecb}

1. "In this task you are presented with multiple documents that are related to the same topic. In each document, each mention of an entity or event is enclosed with square brackets, followed by the mention's unique id. That is: '[some mention](id)'. Your task is to identify and cluster together mentions that are coreferring with each other, i.e., mentions that mark the same entity or event. output the list of clusters, where each cluster is a list of ids of coreferring mentions. For example, if you think that mentions (1) and (2) refer to the same entity, and mentions (3) and (4) refer to the same entity, but these two entities are different, you should output: [[1, 2], [3, 4]]."

2. "Given a set of documents related to the same topic, each containing entities or events marked with unique identifiers in the format '[some mention](id)', your task is to analyze these documents and identify mentions that refer to the same entity or event. Once identified, group these mentions into clusters, where each cluster contains the ids of mentions that are coreferential. Output the clusters in the following format: a list of lists, where each inner list contains the ids of coreferring mentions. For instance, if mentions (1) and (2) are coreferential, and mentions (3) and (4) are coreferential, but refer to different entities, the output should be: [[1, 2], [3, 4]]. Please process the input and produce the correct clusters based on coreference."

3. "Your task is to process a set of documents, each discussing a similar topic and containing marked mentions of entities or events in the format '[some mention](id)'. Identify mentions that refer to the same entity or event and group these ids into clusters. Each cluster should consist of ids that belong to coreferential mentions. Output these clusters in a nested list format, for example, if mentions tagged as (1) and (2) are coreferential, and (3) and (4) are as well, but separate from the first group, your output should look like: [[1, 2], [3, 4]]."

4. "Examine several documents linked by a common theme, noting that each mention of an entity or event within these documents is denoted by '[some mention](id)'. Your objective is to find and group ids of mentions that correspond to the same entity or event into clusters. The desired output format is a list of lists, where each sublist contains ids of mentions that are coreferential. For example, an appropriate output for coreferential mentions (1) and (2), and another pair (3) and (4), would be [[1, 2], [3, 4]]."

5. "Within multiple documents on a similar subject, each entity or event mention is bracketed and tagged with an id in the format '[some mention](id)'. Your role is to ascertain which mentions refer to the same entity or event, grouping these ids into clusters. Each cluster will be a list containing the ids of mentions that are coreferential. For instance, if you determine that (1) and (2), as well as (3) and (4), refer to distinct entities, you should produce the output as: [[1, 2], [3, 4]]."
      
6. "Analyze a series of documents that discuss the same topic where entities and events are noted as '[some mention](id)'. Group mentions that refer to the same entity or event into clusters by their ids. Each cluster should be presented as a list of these ids, such as [[1, 2], [3, 4]], indicating coreference between the mentions within each list."

7. "Review a collection of documents with related topics, identifying and clustering ids of entity and event mentions, denoted as '[some mention](id)'. Mentions that are coreferential should be grouped together. Format your output as a series of lists, each containing ids of mentions that reference the same entity or event, e.g., [[1, 2], [3, 4]]"

 8. "Given multiple documents related by a common theme, each containing marked mentions of entities or events as '[some mention](id)', identify which mentions refer to the same entity or event and cluster these mentions' ids together. Your objective is to find and group ids of mentions that correspond to the same entity or event into clusters. The desired output format is a list of lists, where each sublist contains ids of mentions that are coreferential. For example, an appropriate output for coreferential mentions (1) and (2), and another pair (3) and (4), would be [[1, 2], [3, 4]]."

      9. "Given a collection of documents on the same topic, identify mentions of entities or events marked as '[some mention](id)'. Your task is to group ids of mentions that refer to the same entity or event into clusters. Output these clusters as lists of ids, each representing a set of coreferential mentions. For example, if you find that mentions (1) and (2) refer to the same entity, and (3) and (4) refer to another, output should be: [[1, 2], [3, 4]]."

     10. "Analyze a set of documents that discuss a unified topic, where each entity or event is uniquely identified by '[some mention](id)'. Group together the ids of mentions that are coreferential. Present your findings as a list of lists, where each sublist contains ids of mentions referring to the same entity or event, e.g., [[1, 2], [3, 4]]. Ensure accuracy in identifying and clustering these mentions."

11. "Review documents related by a common theme, each containing marked mentions of entities or events in the format '[some mention](id)'. Determine which mentions refer to the same entity or event and cluster their ids accordingly. Output these clusters as nested lists, where each list contains ids of coreferential mentions, such as [[1, 2], [3, 4]]. This task requires precise identification and clustering based on coreference."

      12. "Your challenge is to navigate through multiple documents related by a common topic, identifying mentions of entities or events enclosed as '[some mention](id)'. Group these mentions by ids when they refer to the same entity or event. Present your clusters as lists of ids that show coreference, for instance, [[1, 2], [3, 4]]. This task requires keen attention to detail and accuracy in identifying connections"
      
      13. "Act as a data detective and delve into several documents, each marked by similar themes and containing entities or events tagged as '[some mention](id)'. Your mission is to uncover which mentions are coreferential and cluster their ids accordingly. Your findings should be reported as lists of ids grouped by coreference, like [[1, 2], [3, 4]]."
     
     14.  "Task: Analyze a collection of thematic documents to identify and cluster mentions of entities or events, formatted as '[some mention](id)', that refer to the same concept. Your output should clearly display clusters of ids representing coreferential mentions, such as [[1, 2], [3, 4]], demonstrating your ability to discern and link related information."
      
      15. "In this detailed analysis task, you are provided with multiple documents, each discussing a similar theme. Within these documents, mentions of entities or events are specifically highlighted and labeled with unique identifiers in the format 'some mention](id)'. Your primary objective is to meticulously identify which of these mentions refer to the same real-world entity or event, and then systematically group together their corresponding ids into clusters. Each cluster should exclusively contain ids of mentions that are coreferential. For clarity in your output, format your results into a list of these clusters, where each sublist represents a distinct group of coreferential mentions. For instance, if your analysis concludes that mentions (1) and (2) are about the same entity, and mentions (3) and (4) about another distinct entity, your output should be formatted as: [[1, 2], [3, 4]]. This task requires precise attention to detail and analytical rigor to ensure accuracy in the clustering process."
      
      16. "Examine several documents on a unified topic, noting each mention of an entity or event as '[some mention](id)'. Group mentions by ids that refer to the same entity or event. Output these clusters in a format of nested lists, with each list containing ids of coreferential mentions, like [[1, 2], [3, 4]]."
      
      17. "In this comprehensive task, you are faced with an array of documents tied by a common thematic element. Each document features various mentions of entities or events, each enclosed within square brackets and followed by a unique identification number in the format '[some mention](id)'. It is your responsibility to sift through these mentions, discerning which ones are references to the same entity or event across different texts. Upon identifying these coreferential mentions, you are to organize and cluster their ids into coherent groups. The final output should be a structured list of these clusters, with each cluster formatted as a list containing ids of mentions that you have determined to be about the same entity or event. For example, if you deduce that mentions (1) and (2) discuss one entity, and mentions (3) and (4) another, your result should be presented as: [[1, 2], [3, 4]]. This task demands a high level of accuracy and a methodical approach to ensure that each cluster is correctly assembled based on coreference analysis."
      
      18. "Within a group of related documents, each mention of an entity or event is uniquely identified in the format '[some mention](id)'. Your objective is to examine these documents and cluster the ids of mentions that are coreferential, i.e., refer to the same entity or event. The output should be a list of lists, where each sublist contains the ids of mentions that are grouped together based on coreference. For example, if mentions (1) and (2) are about the same entity, and (3) and (4) are about another, then the output should look like: [[1, 2], [3, 4]]."
      
      19. "Engage in a task where you need to sift through several documents that deal with similar topics, noting that each entity or event mention is tagged in the format '[some mention](id)'. Identify and group mentions that refer to the same entity or event into clusters by their ids. Produce an output that consists of lists, where each list is a cluster of ids representing coreferential mentions. For instance, if (1) and (2) are deemed coreferential, as are (3) and (4), then your output should be structured as: [[1, 2], [3, 4]]."
       
       20. "Your task involves processing a set of documents on a common subject, where each document includes mentions of entities or events marked by '[some mention](id)'. Your role is to determine which mentions refer to the same entity or event and cluster their ids together. The expected output format is a series of nested lists, with each list containing ids of coreferential mentions. For example, if mentions (1) and (2) are about the same entity, and (3) and (4) are about another, organize your output as: [[1, 2], [3, 4]]."


\end{document}